\documentclass[10pt]{article}
\usepackage[top=3cm,bottom=2cm,left=2cm,right=2cm]{geometry}

\usepackage{float}
\usepackage{cite}
\usepackage{amsthm}
\usepackage{mathtools}
\usepackage{amssymb}
\usepackage{listings}
\usepackage{mathrsfs}
\usepackage{setspace}
\usepackage{graphicx}
\graphicspath{{FIG}}
\usepackage{amsmath}
\usepackage[utf8]{inputenc}
\usepackage[T1]{fontenc}
\usepackage[version=4]{mhchem}
\usepackage{graphicx} 
\usepackage{booktabs}

\usepackage{mathptmx} 
\usepackage[scaled]{helvet} 
\usepackage{courier} 

\usepackage[colorlinks,
            linkcolor=red,
            anchorcolor=blue,
            citecolor=green
            ]{hyperref}

\newtheorem{theorem}{Theorem}[section]

\newcommand{\newcite}[1]{\textnormal{\cite{#1}}}

\begin{document}

\title{An Efficient Approach to Regression Problems with Tensor Neural Networks\footnote{This work 
was supported by the National Key Research and Development Program 
of China (2023YFB3309104), National Natural Science Foundations of 
China (NSFC 1233000214),  National Key Laboratory of Computational Physics 
(No. 6142A05230501),  Beijing Natural Science Foundation (Z200003), 
National Center for Mathematics and Interdisciplinary Science, CAS.}}

\author{Yongxin Li\footnote{School of Statistics and Mathematics, 
Central University of Finance and Economics, 
No.39, Xueyuan Nanlu, Beijing, 100081, China (2018110075@email.cufe.edu.cn)},\ \ \ 
Yifan Wang\footnote{LSEC,
Academy of Mathematics and Systems Science,
Chinese Academy of
Sciences, No.55, Zhongguancun Donglu, Beijing 100190, China, and School of
Mathematical Sciences, University of Chinese Academy
of Sciences, Beijing, 100049 (wangyifan@lsec.cc.ac.cn)},\ \ \
Zhongshuo Lin\footnote{LSEC,
Academy of Mathematics and Systems Science,
Chinese Academy of
Sciences, No.55, Zhongguancun Donglu, Beijing 100190, China, and School of
Mathematical Sciences, University of Chinese Academy
of Sciences, Beijing, 100049 (linzhongshuo@lsec.cc.ac.cn)}\ \ \
and \ \ Hehu Xie\footnote{LSEC,
Academy of Mathematics and Systems Science,
Chinese Academy of
Sciences, No.55, Zhongguancun Donglu, Beijing 100190, China, and School of
Mathematical Sciences, University of Chinese Academy
of Sciences, Beijing, 100049 (hhxie@lsec.cc.ac.cn)}}

\date{}

\maketitle

\begin{abstract}

This paper introduces a tensor neural network (TNN) to address nonparametric regression problems, leveraging its distinct sub-network structure to effectively facilitate variable separation and enhance the approximation of complex, high-dimensional functions. The TNN demonstrates superior performance compared to conventional Feed-Forward Networks (FFN) and Radial Basis Function Networks (RBN) in terms of both approximation accuracy and generalization capacity, even with a comparable number of parameters. A significant innovation in our approach is the integration of statistical regression and numerical integration within the TNN framework. This allows for efficient computation of high-dimensional integrals associated with the regression function and provides detailed insights into the underlying data structure. Furthermore, we employ gradient and Laplacian analysis on the regression outputs to identify key dimensions influencing the predictions, thereby guiding the design of subsequent experiments. These advancements make TNN a powerful tool for applications requiring precise high-dimensional data analysis and predictive modeling.


\end{abstract}

\large
\addtolength{\parskip}{12pt}
\section{Introduction}
Regression analysis is a widely utilized statistical method across various domains, aimed at predicting or modeling the relationship between independent and dependent variables \cite{Hastie2009}. Numerous regression techniques have been developed to accommodate data of diverse scales and characteristics, yielding favorable practical outcomes \cite{Draper1998, montgomery2021}. Despite the successes achieved, there remains a continuous effort to develop more efficient algorithms that enhance both accuracy and interpretability.

As technological advancements across various industries lead to increasingly complex, high-dimensional, and structured datasets, traditional regression methods encounter limitations. These datasets often comprise information from diverse domains such as spatial, imagery, and spectral data, which necessitates the development of models capable of handling such complex structures. Artificial Neural Networks (ANNs), as a widely adopted machine learning algorithm, have been extensively researched and have demonstrated high efficacy in these tasks \cite{Rosenblatt1958, schmidhuber2022}.

Recent developments in deep learning suggest that larger models tend to perform better, leading to the training of networks with hundreds of billions of parameters \cite{Nakkiran2020, kaplan2020, lepikhin2020}. These networks operate in an overparameterized regime, with more parameters than training samples, and can even fit random noise \cite{zhang2021, Belkin2019, Nakkiran2020}. Classical statistical learning theory posits that overparameterization should result in overfitting and degraded generalization \cite{Hastie2009, Belkin2019, Vapnik1998}. However, it has been observed that these networks often generalize well, suggesting that overparameterized neural networks capture essential characteristics of real systems, closely emulating their operations.

Initially, the design of artificial neural networks was inspired by biological neural networks. In this paper, we introduce a neural network architecture inspired by the Candecomp/Parafac (CP) tensor decomposition method, a prevalent low-rank approximation technique \cite{Hitchcock1927, Carroll1970}. This tensor neural network (TNN) was designed for solving high-dimensional PDEs \cite{wang2023tensor, li2024tensor}. We extend the application of TNN to regression problems, demonstrating its superior approximation performance and training convergence with limited data.

Conventional nonlinear regression techniques require a prior specification of the regression equation's form \cite{Hastie2009, Draper1998, montgomery2021}. With a predetermined or guessed equation form, the problem reduces to estimating a small number of constants. However, this approach limits the regression to a "best fit" for the specified equation form, which, if incorrect or unsuitable for the dataset, can lead to significant issues.

Artificial Neural Networks (ANNs), as a nonparametric method, have become extensively studied and effectively applied tools for regression and function approximation tasks. Unlike nonparametric methods that rely on piecewise approximations \cite{Györfi2002, Alexandre2008}, deep architecture ANNs can achieve near-minimax rates for arbitrary smoothness levels of the regression function \cite{KURKOVA1992501, Chen1995, Hieber2020}. ANNs consist of numerous artificial neurons organized into multiple layers, connected in various ways, resulting in different ANN structures. The most common types for regression problems are the Feed-Forward Network (FFN) \cite{Rumelhart1988} and the Radial Basis Network (RBN) \cite{broomhead1988}. 

The success of multilayer neural networks is partly due to efficient algorithms for estimating network weights from data. However, in regression contexts, FFNs often converge to local optima, leading to suboptimal results. The iterative back-propagation algorithms, which minimize an empirical loss function using gradient descent, face challenges due to the nonconvex nature of the function space, such as encountering saddle points or converging to one of many potential local minima \cite{Huang2003, reed1999, looney1997}. Another limitation of FFNs is the modest nonlinearity of multilayer perceptrons, caused by common activation functions like sigmoid, tanh, and ReLU. Attempts to improve accuracy in highly nonlinear problems by adding more layers and neurons often result in increased network complexity and training costs.

In contrast, Radial Basis Transfer Functions used in RBNs make them better suited for nonlinear regression problems \cite{Bianchini1995, Poggio1990}. RBN's convex loss function regarding all weights allows training without local minima issues \cite{Haykin2007}.  However, the design of RBNs often requires heuristic parameter selection based on empirical knowledge, which may yield suboptimal results \cite{maren2014handbook}. The determination of these parameters involves global computations, often utilizing the K-means clustering algorithm \cite{haykin2009neural} and the orthogonal least squares (OLS) algorithm \cite{chen1991} to set RBN basis-function centers. To leverage the strengths of both FFNs and RBNs, deep networks with more than one hidden layer have been developed \cite{Jiang2022, Chao2001}.

The structure of this paper is outlined as follows. In Section 2, we delve into the realm of nonparametric regression problems and explore feed-forward networks, with a particular emphasis on our proposed Tensor Neural Network (TNN). This section covers various aspects of the TNN, including its subnetwork structure, the enhancement of predictive modeling through integration techniques, techniques for input normalization, and strategies for fine-tuning the output approximation. In this section, we also introduce our approach to computing the gradient and Laplacian of the regression function using the TNN framework. These derivatives are crucial for understanding the sensitivity of the predicted outputs to the input variables, which can guide further experimental designs. Section 3 is dedicated to applying the TNN to three illustrative problems to demonstrate its efficacy in comparison with Feed-Forward Networks (FFN) and Radial Basis Networks (RBN). This includes the regression analysis of two multidimensional functions and a complex problem involving eight inputs and one output, as originally described in  \cite{YEH1998} and previously approached using Artificial Neural Networks (ANNs). Finally, Section 4 concludes the paper by summarizing our findings, discussing the broader applications of our study, and suggesting directions for future research in this exciting domain.

\section{Approach} \label{approach}

\subsection{The Nonparametric Regression Problem}
As a nonparametric regression model, artificial neural networks (ANNs) approximate the underlying statistical structure of data by adjusting their architecture and parameters. The effectiveness of this approximation depends on the dataset's characteristics, its scale, the complexity of the regression problem, and the inductive bias of the selected network. For analysis and application convenience, we establish the following mathematical assumptions.

We assume a causal relationship between the inputs $\boldsymbol{x}^k$ and outputs $y^k$ within the dataset, which can be approximated using neural network methodologies. The dataset is assumed to be from continuous numerical domains, sampled on a discrete grid. After normalizing the data, we articulate the following nonparametric regression model:
\begin{align}
y^k =f(\boldsymbol{x}^k) + \varepsilon^k, \quad k =1, \ldots, n.
\label{eq.mod}
\end{align}

Here, $\boldsymbol{x}^k$ are random covariates within the unit hypercube, $\boldsymbol{x}^k \in [0,1]^{d}$, and we observe $n$ independent and identically distributed (i.i.d.) vectors $\boldsymbol{x}^k$ and responses $y^k \in [-1,1]$ from the model, with noisy measurements $\varepsilon^k$.

\subsection{The Feed-forward Networks}

\begin{figure}[H]
\centering
\includegraphics{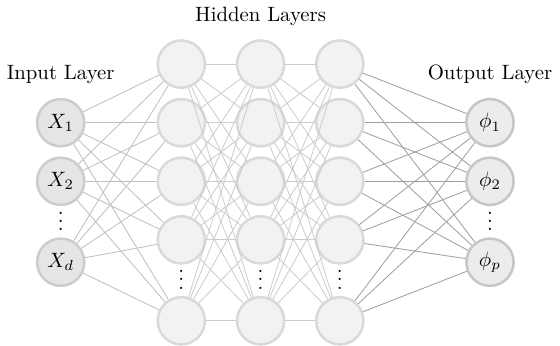}
\caption{Schematic of a feed-forward network, showing data propagation from the Input Layer to the Sum Layer.}
\label{fig:FFNArc}
\end{figure}

Feed-forward neural networks (FFNs) are fundamental architectures in machine learning and deep learning. Characterized by a unidirectional flow of information from the input layer through to the output layer, FFNs lack feedback loops. Neurons in each layer are fully connected to those in adjacent layers, facilitating linear or affine transformations, as depicted in Figure \ref{fig:FFNArc}.

The FFN's architecture, denoted as $(L, \boldsymbol{p})$, comprises a positive integer $L$ (the number of hidden layers or depth) and a width vector $\boldsymbol{p}=(p_0, \ldots, p_{L+1}) \in \mathbb{N}^{L+2}.$ Here, $p_0$ matches the input dimensionality $d$, and $p_{L+1}$ is the output layer width. The FFN, represented as $\boldsymbol{\Phi}(\boldsymbol{x};\theta): \mathbb{R}^{p_0} \rightarrow \mathbb{R}^{p_{L+1}}$, is defined as:
\begin{eqnarray}\label{def_FFN}
\boldsymbol{\Phi}(\boldsymbol{x};\theta)=\big(\phi_{1}(\boldsymbol{x};\theta),\ldots,\phi_{p}(\boldsymbol{x};\theta)\big)^T,
\end{eqnarray}
where $\theta$ denotes all the network parameters, typically weights and biases. The network $\boldsymbol{\Phi}(\boldsymbol{x};\theta)$ with architecture $(L, \boldsymbol{p})$ is any function of the form:
\begin{equation}
\left\{\begin{aligned}
&\boldsymbol{y}_{0}=\boldsymbol{x}, \\
&\boldsymbol{y}_{l} =  \sigma_l(\boldsymbol{W}_l \boldsymbol{y}_{l-1}+\boldsymbol{b}_l), \quad l = 1, \dots , L+1, \\
&\boldsymbol{\Phi}(\boldsymbol{x}) = \boldsymbol{y}_{L+1}
\end{aligned}\right.
\end{equation}
with $\boldsymbol{W}_l$ as a $p_l\times p_{l-1}$ weight matrix and $\boldsymbol{b}_l \in \mathbb{R}^{p_l}$ as the bias vector. The network functions are built by alternating matrix-vector multiplications with the action of non-linear activation functions $\sigma_l, l = 1, \ldots, L+1.$

In employing FFN for the regression problem outlined in Equation (\ref{eq.mod}), we set $p = 1$ and proceed to train the neural network using the sample data.

\subsection{The Architecture of Tensor Neural Networks}

\begin{figure}[H]
\centering
\includegraphics{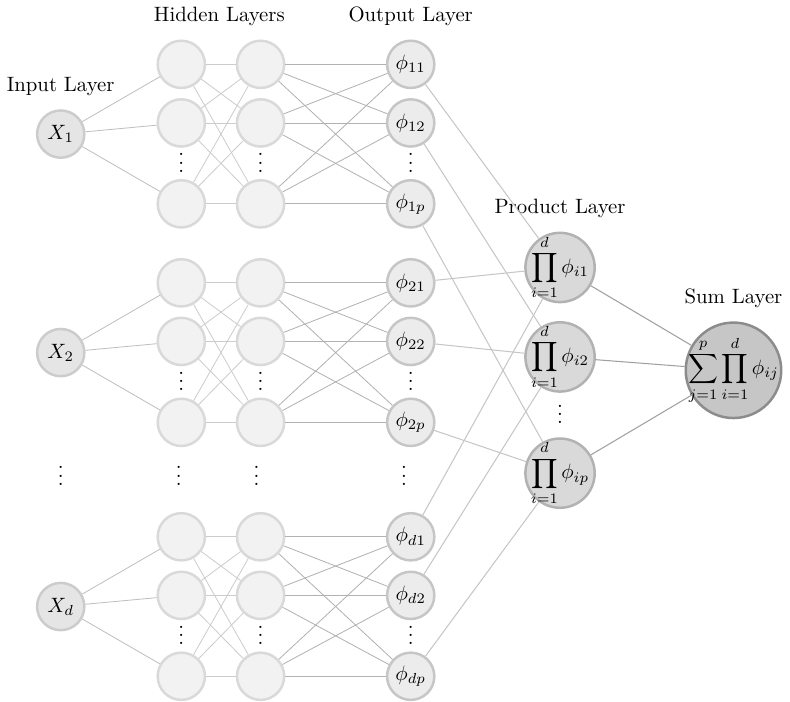}
\caption{Schematic of the tensor neural network, illustrating unidirectional data propagation from the Input Layer to the Sum Layer.}
\label{fig:NetArc}
\end{figure}

The Tensor Neural Network (TNN) is inspired by two theoretical achievements: the Singular Value Decomposition (SVD) of polynomials, a powerful tool in mathematical analysis \cite{Eckart1936}, and the universal approximation theorem, which emphasizes the expressive power of neural networks \cite{KURKOVA1992501, Chen1995}. The TNN's architecture, as depicted in Figure \ref{fig:NetArc}, consists of $d$ subnetworks corresponding to the input data's dimensionality. Each subnetwork uses a 1-dimensional input and a $p$-dimensional output FFN, converging to a one-dimensional output after product and sum layers.

The TNN architecture is formalized as:
\begin{eqnarray}\label{def_TNN}
\Psi(\boldsymbol{x};\theta)=\sum_{j=1}^p\prod_{i=1}^d\phi_{i,j}(x_i;\theta_i),
\end{eqnarray}
where $\boldsymbol{x} \in \mathbb{R}^{d}$ represents the input vector in a $d$-dimensional real space, $\theta = \{\theta_1, \ldots, \theta_d\}$ denotes all the parameters of the TNN, and $\theta_i$ represents the parameters of the $i$-th subnetwork.

Each subnetwork, $\boldsymbol{\Phi}_i(x_i;\theta_i)$, maps $\mathbb R$ to $\mathbb R^p$ as follows:
\begin{eqnarray}\label{def_FNN}
\boldsymbol{\Phi}_i(x_i;\theta_i)=\big(\phi_{i,1}(x_i;\theta_i),\ldots,\phi_{i,p}(x_i;\theta_i)\big)^T,\ \ \ i=1, \ldots, d.
\end{eqnarray}
The subnetwork operates by alternating matrix-vector multiplications with non-linear activation functions. The choice of activation functions and the network architecture, including the number of layers and neurons, can vary across subnetworks, provided their output dimensions are the same.

To fit data from the $d$-variate nonparametric regression model, we set $p_0=1$ and $p_{L+1}=p$. The TNN, denoted as $\Psi(\boldsymbol{x};\theta)$, provides a sparse tensor approximation of the regression function $f(\boldsymbol{x})$. Network structure optimization and parameter tuning are crucial for closely approximating the regression function.

\subsection{Integration and Prediction}

In practical applications, it is widely recognized that the integration of the high-dimensional FFN functions is typically constrained to Monte-Carlo methods, known for their lower accuracy.  In \cite{wang2023tensor}, an elegant quadrature scheme is introduced in conjunction with TNNs to decompose high-dimensional integration $\int_\Omega f(x)dx$ into a series of 1-dimensional integrations. This scheme can reduce the computational work of high-dimensional integration for the $d$-dimensional function $f(x)$ to a polynomial scale with respect to dimension $d$.

For $\boldsymbol{\Omega} =\Omega_1\times\Omega_2\times\cdots \times\Omega_d  \in \mathbb{R}^{d}$, the TNN integration can be expressed as:
\begin{align}\label{Int_TNN}
\int_{\boldsymbol{\Omega}} d\boldsymbol{x} \Psi(\boldsymbol{x};\theta) 
&=\int_{\boldsymbol{\Omega}} d\boldsymbol{x} \sum_{j=1}^p\prod_{i=1}^d\phi_{i,j}(x_i;\theta_i) \notag \\
&=\int_{\Omega_1\times\Omega_2\times\cdots \times\Omega_d } dx_1 dx_2 \cdots dx_d \sum_{j=1}^p\phi_{1,j}(x_1;\theta_1)\phi_{2,j}(x_2;\theta_2)\cdots\phi_{d,j}(x_d;\theta_d) \\
&=\sum_{j=1}^p\prod_{i=1}^d \int_{\Omega_i} dx_i \phi_{i,j}(x_i;\theta_i) \notag,
\end{align}
and,
\begin{align}\label{Int_TNN2}
\int_{\boldsymbol{\Omega}} d\boldsymbol{x} \Psi^2(\boldsymbol{x};\theta) 
&=\int_{\boldsymbol{\Omega}} d\boldsymbol{x} (\sum_{j=1}^p\prod_{i=1}^d\phi_{i,j}(x_i;\theta_i))^2 \notag \\
&=\int_{\Omega_1\times\Omega_2\times\cdots \times\Omega_d } dx_1 dx_2 \cdots dx_d  \sum_{j=1}^p \sum_{k=1}^p \prod_{i=1}^d \phi_{i,j}(x_i;\theta_i)\phi_{i,k}(x_i;\theta_i) \\
&=\sum_{j=1}^p\sum_{k=1}^p\prod_{i=1}^d \int_{\Omega_i} dx_i  \phi_{i,j}(x_i;\theta_i)\phi_{i,k}(x_i;\theta_i) \notag,
\end{align}

The core contribution of our research is the adaptation of Tensor Neural Networks (TNNs) for regression tasks, which facilitates the extraction of detailed information from regression functions. This adaptation capitalizes on the inherent strength of TNNs in managing high-dimensional integrations, a capability we leverage to uncover deeper insights into the underlying patterns within the regression data.

In this study, we utilize TNNs with Mean Squared Error (MSE) as the loss function, transforming them into potent tools for interpolation. A significant innovation of our approach lies in employing TNNs for high-precision interpolation, which plays a crucial role in accurately estimating function values at points where explicit data is unavailable—an essential step in complex computational tasks.

To compute high-dimensional integrals of functions interpolated by TNNs, we employ the Gaussian integration method. Known for its high precision in one-dimensional contexts, this method aligns seamlessly with the dimensional reduction capabilities intrinsic to TNNs. This synergistic combination enables us to perform high-dimensional integrations with improved accuracy and efficiency.

Let's consider each dimension $i$, ranging from $1$ to $d$. For these dimensions, we select $N_i$ Gaussian points, represented by $\{x_i^{(n_i)}\}_{n_i=1}^{N_i}$, along with their respective weights  $\{w_i^{(n_i)}\}_{n_i=1}^{N_i}$, for the corresponding dimensional domain $\Omega_i$. Here, $N$ symbolizes the maximum number of Gaussian points across all dimensions, defined as $N=\max\{N_1,\cdots,N_d\}$, while $\underline N$ represents the minimum, stated as and
$\underline N = \min\{N_1,\cdots,N_d\}$. By defining the index $n=(n_1,\cdots,n_d)$ within the set $\mathcal N\coloneqq\{1,\cdots,N_1\}\times\cdots\times\{1,\cdots,N_d\}$, we can systematically outline the Gaussian points and weights used in the integrations specified by Equations (\ref{Int_TNN}) and (\ref{Int_TNN2}). These are presented as follows:
\begin{eqnarray}\label{def_Gauss}
\left.
\begin{array}{rcl}
\Big\{x^{(n)}\Big\}_{n\in\mathcal N}&=&\left\{\left\{x_1^{(n_1)}\right\}_{n_1=1}^{N_1},\ \left\{x_2^{(n_2)}\right\}_{n_2=1}^{N_2},\ \cdots,\ \left\{x_d^{(n_d)}\right\}_{n_d=1}^{N_d}\right\},\\
\Big\{w^{(n)}\Big\}_{n\in\mathcal N}&=&\left\{\left\{w_1^{(n_1)}\right\}_{n_1=1}^{N_1}\times \left\{w_2^{(n_2)}\right\}_{n_2=1}^{N_2}\times \cdots, \times  \left\{w_d^{(n_d)}\right\}_{n_d=1}^{N_d}\right\}.
\end{array}
\right.
\end{eqnarray}

Then, the numerical integration (\ref{Int_TNN}) and (\ref{Int_TNN2})
can be computed as follows:

\begin{align}\label{Int_TNN_G}
\int_{\boldsymbol{\Omega}} d\boldsymbol{x} \Psi(\boldsymbol{x};\theta) 
\approx \sum_{j=1}^p\prod_{i=1}^d \left(\sum_{n_i=1}^{N_i}w_i^{(n_i)} \phi_{i,j}(x_i^{(n_i)};\theta_i) \right),
\end{align}
and,
\begin{align}\label{Int_TNN2_G}
\int_{\boldsymbol{\Omega}} d\boldsymbol{x} \Psi^2(\boldsymbol{x};\theta) 
\approx \sum_{j=1}^p\sum_{k=1}^p\prod_{i=1}^d \left(\sum_{n_i=1}^{N_i}w_i^{(n_i)}  \phi_{i,j}(x_i^{(n_i)};\theta_i)\phi_{i,k}(x_i^{(n_i)};\theta_i)\right),
\end{align}

Theorem 3 of \newcite{wang2023tensor} gives the corresponding result. Since the 1-dimensional integration with $N_i$ Gauss points has $2N_i$-th order accuracy and the equivalence of (\ref{Int_TNN_G}) and (\ref{Int_TNN2_G}), both quadrature schemes (\ref{Int_TNN_G}) and (\ref{Int_TNN2_G}) have the $2\underline N$-th order accuracy. 

Now, let us consider $\boldsymbol{x_{0}}$ as the data point of interest, where its integration dimension and range, along with the corresponding statistical information, are governed by $\rho_{\boldsymbol{x_{0}}}(\boldsymbol{x})=\prod_{i=1}^d\rho_{\boldsymbol{x_{0}}i}(x_i)$. Namely, $\rho_{\boldsymbol{x_0}}$ represents the integration weight function determined by $\boldsymbol{x_0}$. Our final solution, denoted as $f(\boldsymbol{x_{0}})$, is given by Equation (\ref{Int_TNN_G_f}) as follows:

\begin{align}\label{Int_TNN_G_f}
f(\boldsymbol{x_{0}}) & \coloneqq \frac{\int_{\boldsymbol{\Omega}} d\boldsymbol{x} \Psi(\boldsymbol{x};\theta) \rho_{\boldsymbol{x_{0}}i}(\boldsymbol{x}) }{\int_{\boldsymbol{\Omega}} d\boldsymbol{x} \rho_{\boldsymbol{x_{0}}i}(\boldsymbol{x})} \nonumber \\
& \approx \frac{\sum_{j=1}^p\prod_{i=1}^d \left(\sum_{n_i=1}^{N_i}w_i^{(n_i)} \phi_{i,j}(x_i^{(n_i)};\theta_i) \rho_{\boldsymbol{x_{0}}i}(x_i^{(n_i)})\right)}{ \sum_{j=1}^p\prod_{i=1}^d \left(\sum_{n_i=1}^{N_i}w_i^{(n_i)} \rho_{\boldsymbol{x_{0}}i}(x_i^{(n_i)})\right)},
\end{align}

Similarly, the prediction of $f^2(\boldsymbol{x_{0}})$ is provided by Equation (\ref{Int_TNN2_G_f}):

\begin{align}\label{Int_TNN2_G_f}
f^2(\boldsymbol{x_{0}}) & \coloneqq \frac{\int_{\boldsymbol{\Omega}} d\boldsymbol{x} \Psi^2 (\boldsymbol{x};\theta) \rho_{\boldsymbol{x_{0}}i}(\boldsymbol{x}) }{\int_{\boldsymbol{\Omega}} d\boldsymbol{x} \rho_{\boldsymbol{x_{0}}i}(\boldsymbol{x})} \nonumber \\
& \approx \frac{\sum_{j=1}^p\sum_{k=1}^p\prod_{i=1}^d \left(\sum_{n_i=1}^{N_i}w_i^{(n_i)} \phi_{i,j}(x_i^{(n_i)};\theta_i)\phi_{i,k}(x_i^{(n_i)};\theta_i)\rho_{\boldsymbol{x_{0}}i}(x_i^{(n_i)}) \right)}{\sum_{j=1}^p\sum_{k=1}^p\prod_{i=1}^d \left(\sum_{n_i=1}^{N_i}w_i^{(n_i)} \rho_{\boldsymbol{x_{0}}i}(x_i^{(n_i)}) \right)},
\end{align}

\subsection{Gradient and Laplacian Analysis for Experimental Design}

In addition to deriving regression functions, Tensor Neural Networks (TNNs) offer a powerful means of extracting gradients and Laplacians at specific data points. This capability significantly enhances our ability to guide experimental processes by providing deeper insights into the behavior of underlying functions across various regions of the input space. By analyzing these derivatives, we gain a more nuanced understanding of the system under study, which in turn informs and optimizes experimental design.

In the context of Tensor Neural Networks (TNNs), the computation of gradients and the Laplacian operator for high-dimensional input functions is crucial for enhancing the understanding and interpretation of the modeled system. Specifically, given a TNN output function \(\Psi(\boldsymbol{x}; \theta)\), where \(\boldsymbol{x} = [x_1, x_2, \dots, x_d]\) represents the input variables and \(\theta\) denotes the TNN parameters, the gradient \(\nabla \Psi(\boldsymbol{x}; \theta)\) and the Laplacian \(\Delta \Psi(\boldsymbol{x}; \theta)\) can be efficiently derived.

\subsubsection{Gradient Representation}

The gradient \(\nabla \Psi(\boldsymbol{x}; \theta)\) is expressed as a vector of first-order partial derivatives with respect to each input variable \(x_i\):

\begin{equation}\label{grad_TNN}
\nabla f(\boldsymbol{x}) \coloneqq \nabla \Psi(\boldsymbol{x}; \theta) = \left[\frac{\partial \Psi(\boldsymbol{x}; \theta)}{\partial x_1}, \frac{\partial \Psi(\boldsymbol{x}; \theta)}{\partial x_2}, \dots, \frac{\partial \Psi(\boldsymbol{x}; \theta)}{\partial x_d}\right]
\end{equation}

Within the TNN framework, this gradient can be decomposed for each dimension as follows:

\begin{equation}
\frac{\partial \Psi(\boldsymbol{x}; \theta)}{\partial x_i} = \sum_{j=1}^{p} \left( \frac{\partial \phi_{i,j}(x_i; \theta_i)}{\partial x_i} \prod_{k \neq i} \phi_{k,j}(x_k; \theta_k) \right)
\end{equation}

Here, \(\phi_{i,j}(x_i; \theta_i)\) represents the output of the sub-network corresponding to the \(i\)th dimension, while \(\frac{\partial \phi_{i,j}(x_i; \theta_i)}{\partial x_i}\) denotes its first-order derivative with respect to \(x_i\).

\subsubsection{Laplacian Representation}

The Laplacian operator, being the divergence of the gradient, is given by the sum of the second-order partial derivatives of the function:

\begin{equation}\label{lap_TNN}
\Delta f(\boldsymbol{x}) \coloneqq  \Delta \Psi(\boldsymbol{x}; \theta) = \sum_{i=1}^{d} \frac{\partial^2 \Psi(\boldsymbol{x}; \theta)}{\partial x_i^2}
\end{equation}

In TNNs, this can be represented as:

\begin{equation}
\frac{\partial^2 \Psi(\boldsymbol{x}; \theta)}{\partial x_i^2} = \sum_{j=1}^{p} \left( \frac{\partial^2 \phi_{i,j}(x_i; \theta_i)}{\partial x_i^2} \prod_{k \neq i} \phi_{k,j}(x_k; \theta_k) \right)
\end{equation}

This approach leverages the separability of variables within TNNs, facilitating straightforward computation of second-order derivatives across multiple dimensions.

\subsection{Normalization of Inputs and Fine-Tuning the Output Approximation}
Normalization is a critical preprocessing step, essential for achieving uniformity in the ranges or variances of all input variables. This process ensures equitable weightings across each dimension of the input data, preventing variables with larger numerical values from disproportionately influencing the prediction process. Moreover, normalization enhances the numerical stability and generality of the model.

In this study, we define the inputs, $\boldsymbol{x}$, as random covariates within the unit hypercube, $\boldsymbol{x} \in [0,1]^{d}$, and set $y \in [-1,1]$. For the observed $n$ i.i.d. vectors $\boldsymbol{X}^i$ and $n$ responses $Y^i$, we apply an affine transformation within each dimension to preserve the data structure:
\begin{equation*}
x_{i}^{m} = \frac{X_i^m - X_{i_{\text{min}}}}{X_{i_{\text{max}}} - X_{i_{\text{min}}}} ,\quad
y^m = \frac{Y^m - Y_{\text{mean}}}{Y_{\text{max}} - Y_{\text{min}}}
\end{equation*}
where i=1,\ldots,d, and m=1,\ldots,n.  Here, ${X_{i_{\text{max}}}, X_{i_{\text{min}}}}$ represent the maximum and minimum values of the independent variable in the $i$-th dimension of the sample data, and $Y_{\text{mean}}, Y_{\text{max}}$ and $Y_{\text{min}}$ denote the mean, maximum, and minimum values of the dependent variable. 
$\boldsymbol{x}^m = (x_1^m, x_2^m,\ldots, x_d^m)$, with $(\boldsymbol{x}^m, y^m)$ as the $m$-th training case.

The following theorem elucidates the function approximation properties of TNN:
\begin{theorem} \newcite{wang2023tensor}\label{theorem_approximation}
Assume that each $\Omega_i$ is a bounded open interval in $\mathbb R$ for $i=1, \cdots, d$, $\Omega=\Omega_1\times\cdots\times\Omega_d$, and the function $f(x)\in H^m(\Omega)$. For any tolerance $\varepsilon>0$, there exist a positive integer $p$ and a corresponding TNN defined by (\ref{def_TNN}) such that:
\begin{equation}\label{eq:L2_app}
|f(x)-\Psi(x;\theta)|_{H^m(\Omega)}<\varepsilon.
\end{equation}

\end{theorem}

Theorem \ref{theorem_approximation} indicates that a TNN can achieve any desired level of approximation accuracy for a function $f(x)\in H^m(\Omega)$ by appropriately tuning the width of the subnetwork output layer.

In addition to subnetwork width, subnetwork depth is another crucial hyperparameter. For nonparametric regression, it has been recommended to scale the network depth with the sample size \cite{Hieber2020}. Using a richer class of composition-based functions $\mathcal{G}(q,\boldsymbol{d},\boldsymbol{t},\boldsymbol{\beta},K)$ and a class of sparse multilayer neural networks $\mathcal{F}(L, \boldsymbol{p}, s, F)$, it is demonstrated that least squares estimation over sparse neural networks (with an appropriately chosen architecture $(L, \boldsymbol{p}, s, F)$) nearly achieves the minimax prediction error over $\mathcal{G}(q,\boldsymbol{d},\boldsymbol{t},\boldsymbol{\beta},K)$. This suggests that, given a suitable subnetwork architecture $(L, \boldsymbol{p})$, $\Psi(\boldsymbol{x};\theta)$ can approximate regression functions with varying smoothness levels.

\subsection{Training of TNN}

This section outlines the optimization procedure, a critical component of the TNN training process. TNNs are implemented using the open-source machine learning framework PyTorch \cite{paszke2017automatic}. The adaptive moment estimation (Adam) optimizer \cite{kingma2017adam}, a first-order, gradient-based algorithm developed from stochastic gradient descent \cite{Bottou2010}, is used for training. The optimization process aims to find networks $\Psi(\cdot;\theta)$ with minimal empirical risk, using a loss function $E$ defined as the mean squared error (MSE) of the predicted output $\Psi(\boldsymbol{x}^k;\theta)$ relative to the target $y^k$:

\begin{equation*}
E(\boldsymbol{x},y; \theta) = \frac{1}{n}\sum_{k=1}^{n}(\Psi(\boldsymbol{x}^k)-y^k)^2.
\end{equation*}

Here, $\boldsymbol{x}$ and $y$ represent the input and output training datasets, respectively, and $\theta$ encompasses all network parameters. During training, the parameters $\theta$ are updated iteratively to minimize the loss function, typically using a gradient descent approach. The parameter update in each iteration step is as follows:

\begin{equation*}
\begin{aligned}
\boldsymbol{W}_{l}  \quad \leftarrow & &\boldsymbol{W}_{l} - \eta \frac{\partial E(\boldsymbol{x},y; \theta)}{\partial \boldsymbol{W}_{l}}\\
\boldsymbol{b}_{l} \quad  \leftarrow  & &\boldsymbol{b}_{l} - \eta \frac{\partial E(\boldsymbol{x},y; \theta)}{\partial \boldsymbol{b}_{l}}, 
\end{aligned}
\end{equation*}

where, $\eta > 0$ represents the learning rate. The gradient descent method involves calculating the partial derivatives of the loss function with respect to the network parameters. In PyTorch, these partial derivatives are computed using automatic differentiation (auto-grad) mechanisms. 

The auto-grad mechanism records all data-generating operations during the forward pass of the neural network, creating a computation graph. This graph has input tensors as leaf nodes and output tensors as the root. Each operation stores local gradient information, which is combined during the backward pass using the chain rule to yield partial derivatives of the loss function. With this mechanism, the loss function is minimized iteratively by updating $\theta$ in each training epoch.

Simultaneously, hyperparameters such as the subnetwork architecture $(L, \boldsymbol{p})$ are meticulously adjusted throughout the training process to optimize regression performance.

\section{Numerical Examples}

Building upon the Tensor Neural Networks (TNNs) introduced in Section \ref{approach}, this section delves into their application in regression problems through three illustrative examples. The first two examples are specifically designed to demonstrate the accuracy of TNNs as a nonparametric regression model. The third example involves a more complex scenario with eight inputs and one output, initially explored in \cite{YEH1998} and addressed using artificial neural networks (ANNs). The dataset for this second example is sourced from \cite{Yeh2007}.

In all the cases, we observe improvements in training convergence and approximation performance, achieved through fine-tuning the subnetworks' depth and width, guided by relevant performance metrics. These examples effectively demonstrate the capability of TNNs to approximate unknown models with high accuracy.

For a regression problem, the entire database is segmented into training, validation, and testing datasets, each comprising both input and output data. During the training phase, the training datasets play a pivotal role in optimizing network parameters to minimize training loss. Concurrently, the validation dataset is utilized to monitor the validation loss, serving as a mechanism for early stopping to prevent overfitting. Upon completion of the training process, the well-trained networks are evaluated for their generalizability using the testing datasets. Unless specified otherwise, the training-validation split ratio is maintained at 9:1.

In the experimental analysis, the performance of TNN is benchmarked against that of the Feed-Forward Neural Network (FFN) and the Radial Basis Network (RBN) \cite{wasserman1993}, all implemented using the PyTorch framework with comparable parameter. The primary evaluation metric employed is the mean squared error (MSE), which provides a quantitative measure of the average magnitude of errors across the test samples. This standard method of error calculation is essential for objectively assessing the accuracy of each model in predicting the outcomes of the given regression problems.

The implemented FFN includes 4 hidden layers, with each layer comprising 40 neurons, an input layer, and a concluding output layer. The RBN, on the other hand, is set up with 80 radial basis function units. Each of these units has its own center and sigma parameters. The network concludes with a linear output layer. The TNN consists of 8 sub-networks. Each sub-network has 3 hidden layers and  a output layer, each layer contains 5 neurons.  Both the FFN and the TNN use the tanh function as their activation function.

To ensure consistency and robustness in our evaluation, all networks are trained under identical conditions as detailed in Table $\ref{table: hyperpara}$. To mitigate the effects of randomness introduced by the initial random parameter initialization, each experimental case is executed 20 times, and the results are then averaged across these runs.

\begin{table}[H]
  \centering
  \caption{Training hyperparameters used for FFN, RBN and TNN}
  \label{table: hyperpara}
 \begin{tabular}{lccc}
 \toprule  
 Hyperparameter	& Value \\
 \midrule  
 Initial Learning Rate	&	0.001	\\
 Learning Rate Reducing Factor 	&	0.5	\\
 Learning Rate Reducing Patience	&	500	\\
 Early Stopping Patience 	&	1000	\\
 Training–Validation Split 	&	9:1	\\
 \bottomrule  
 \end{tabular}
\end{table}

The foundation of the TNN regression method's efficacy is epitomized by Equations (\ref{Int_TNN_G_f}) and (\ref{Int_TNN2_G_f}), which delineate a predictive mechanism based on the trained neural representation $\Psi(\boldsymbol{x})$. These equations illuminate the TNN method's capacity to approximate the integrals of a function $f(\boldsymbol{x})$ and its square $f^2(\boldsymbol{x})$ over a domain $\Omega$, denoted by $\int_{\Omega}d\boldsymbol{x}f(\boldsymbol{x})$ and $\int_{\Omega}d\boldsymbol{x}f^2(\boldsymbol{x})$, respectively, through the integrals involving $\Psi(\boldsymbol{x})$ and $\Psi^2(\boldsymbol{x})$. The precision of this approximation is critical, as it directly influences the method's overall performance in predicting $f(\boldsymbol{x_0})$ and $f^2(\boldsymbol{x_0})$ for a given data point $\boldsymbol{x_0}$, underpinned by the weight function $\rho_{\boldsymbol{x_0}}(\boldsymbol{x})$.

Equation (\ref{Int_TNN_G_f}) presents the methodology to predict the value of $f(\boldsymbol{x_0})$ through an integration over $\Omega$, weighted by $\rho_{\boldsymbol{x_0}}(\boldsymbol{x})$, and approximated using the trained neural network outputs. Similarly, Equation (\ref{Int_TNN2_G_f}) extends this framework to predict $f^2(\boldsymbol{x_0})$, highlighting the TNN method's adaptability in handling both linear and quadratic function predictions.

The TNN method's predictive accuracy was validated through rigorous testing, confirming its ability to closely approximate integral values of $f(\boldsymbol{x})$ and $f^2(\boldsymbol{x})$. This was evidenced by its performance against the actual integrals $\int_{\Omega}d\boldsymbol{x}f(\boldsymbol{x})$ and $\int_{\Omega}d\boldsymbol{x}f^2(\boldsymbol{x})$, showcasing TNN's precision in complex integration tasks. And a network with a specific architecture of $L = 3$ layers and $\boldsymbol{p} = (1, 20, 20, 20, 20)$ was deployed for 8-dimensional space tests.

 \subsection{Example of 8-Dimensional Sum Function}
This section illustrates the application of Tensor Neural Networks (TNNs) in regression problems using a specific example.  The example involves a continuous function $f(\boldsymbol{x})$ defined over an 8-dimensional input vector $\boldsymbol{x}$:
 
\begin{equation*}
f(\boldsymbol{x}) = \sum_{i=1}^8 sin(2 \pi x_i), \quad \boldsymbol{x} \in [0,1]^8.
\end{equation*}

We conducted random sampling to extract 1000 i.i.d. vectors $\boldsymbol{x}^k, k=1,\ldots, 1000,$ from $[0,1]^8$. These vectors were then processed through the function $f(\boldsymbol{x})$ to create a dataset of input-output pairs $(\boldsymbol{x}^k, f(\boldsymbol{x^k})),$ $k=1,\ldots, 1000$. Out of these, 800 pairs were allocated to the training set, and the remaining 200 were reserved for the test set. The Tanh function was employed as the activation function.

\begin{figure}[H]
\centering
\includegraphics[width=7cm,height=6.5cm]{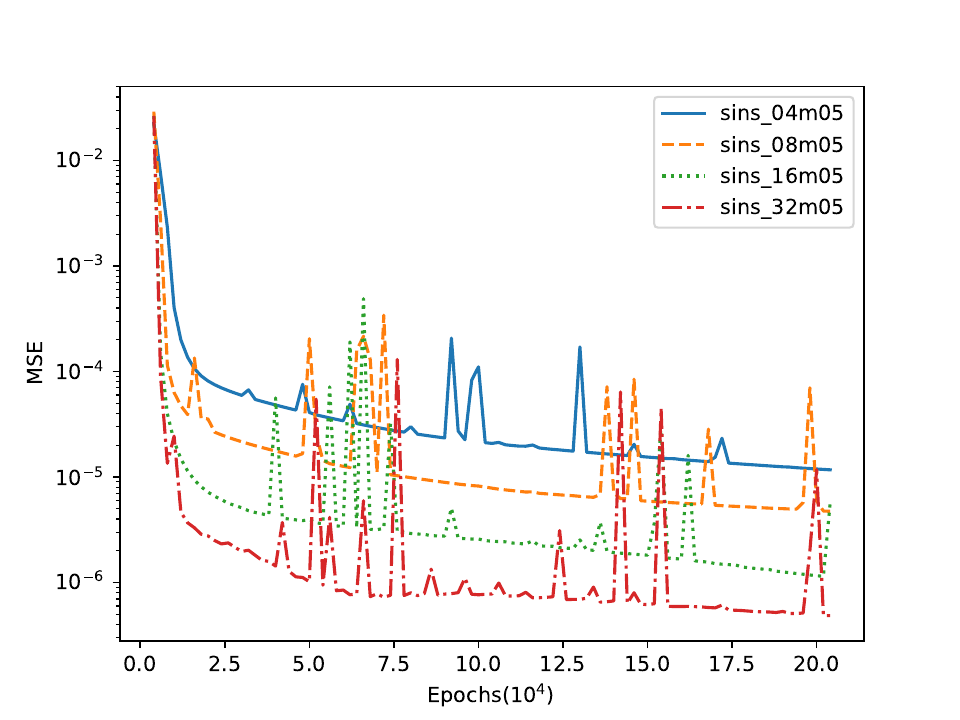}
\includegraphics[width=7cm,height=6.5cm]{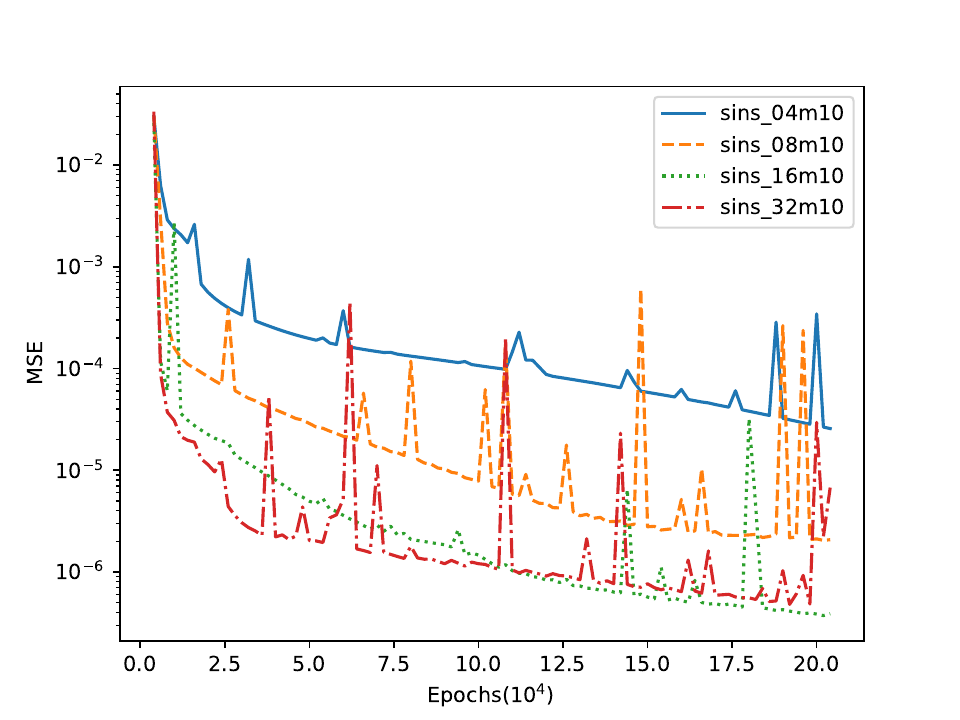}\\
\includegraphics[width=7cm,height=6.5cm]{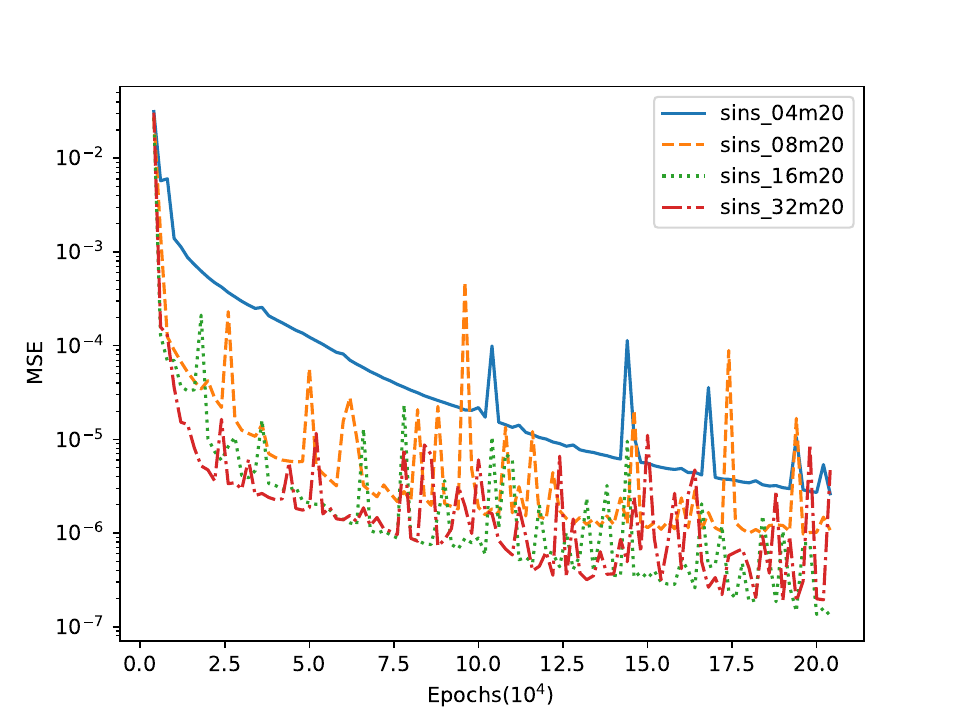}
\includegraphics[width=7cm,height=6.5cm]{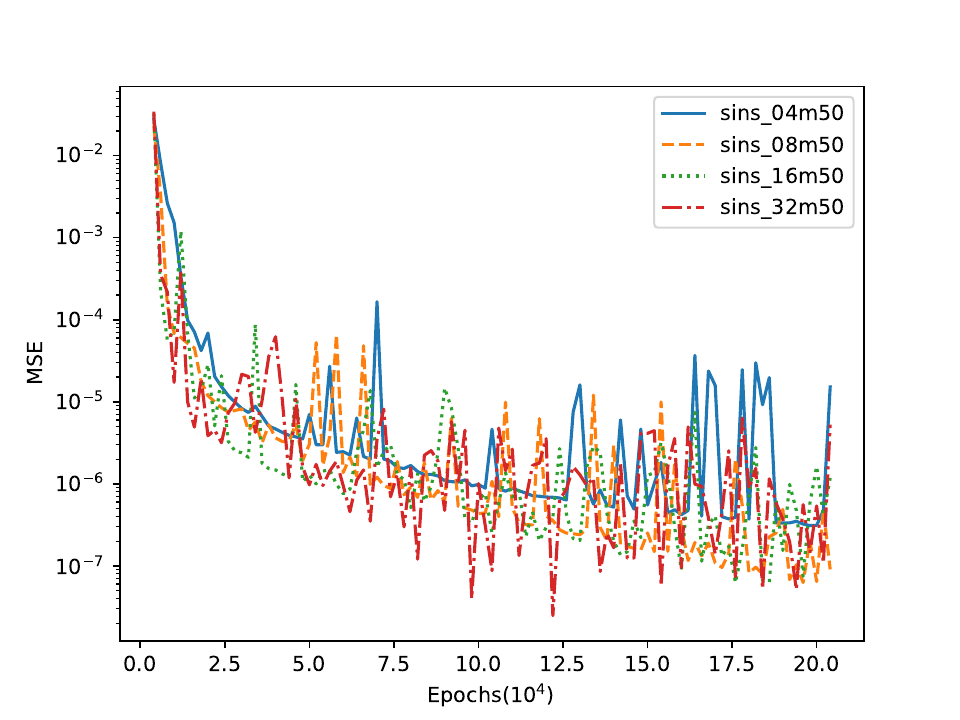}\\
\caption{Training set MSE for different depth settings of subnetworks in the approximation of the continuous function. The convergence speed of the TNN network is influenced by network depth, with increased network width reducing this correlation.}\label{fig_sins_depth}
\end{figure}

\begin{figure}[H]
\centering
\includegraphics[width=7cm,height=6.5cm]{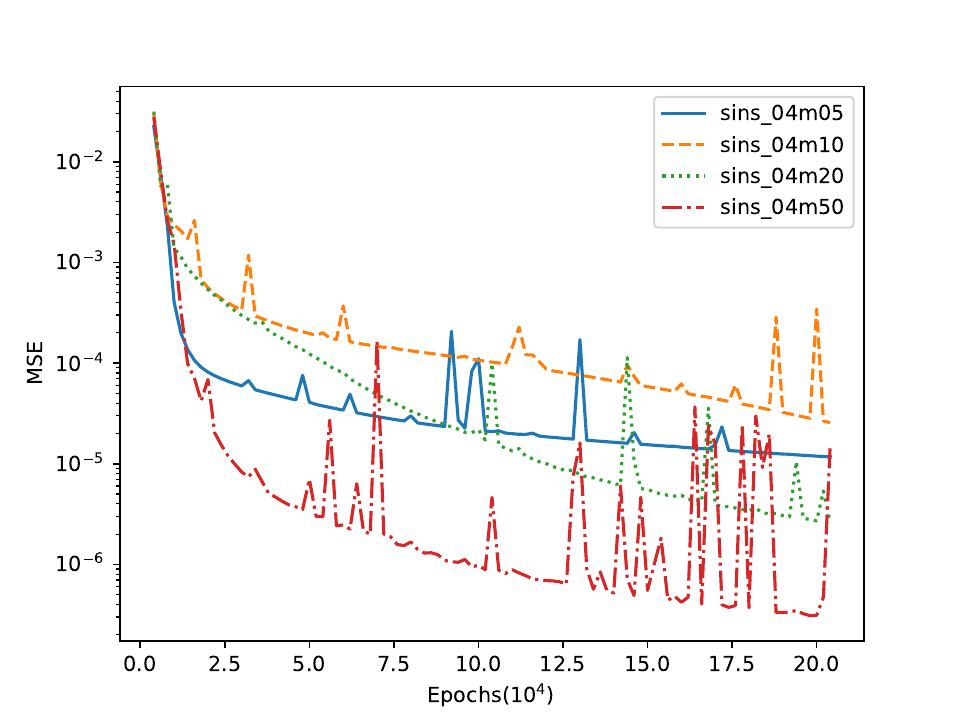}
\includegraphics[width=7cm,height=6.5cm]{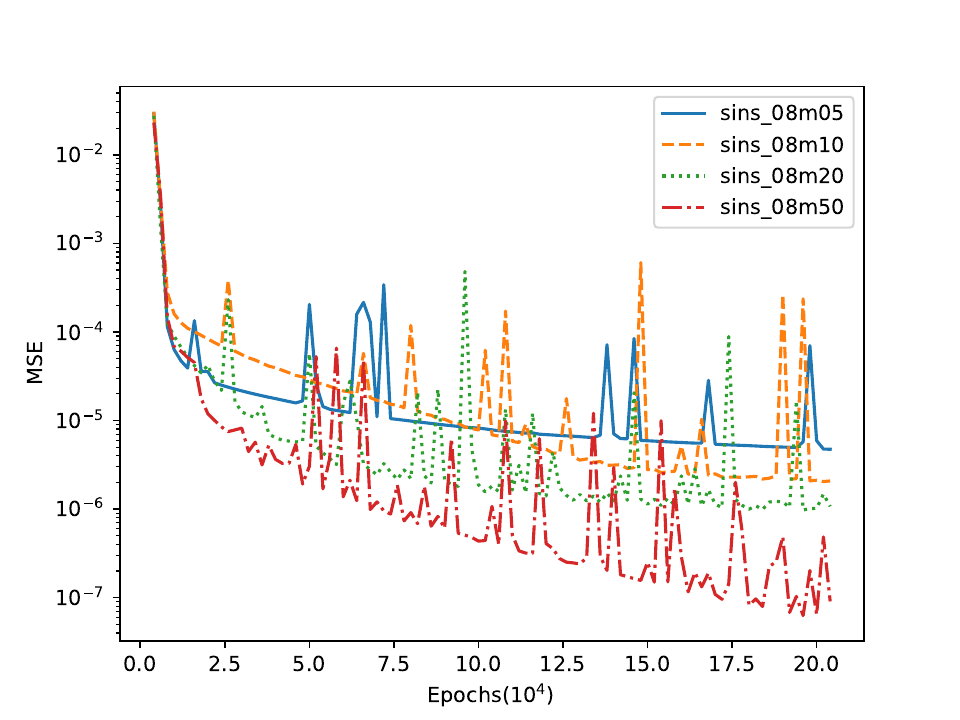}\\
\includegraphics[width=7cm,height=6.5cm]{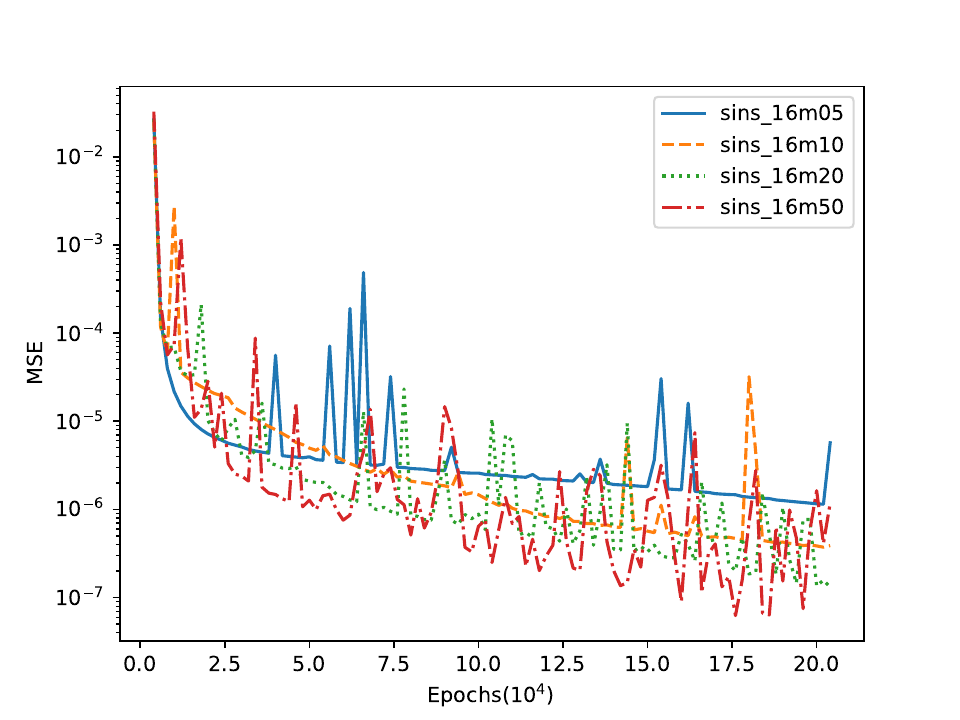}
\includegraphics[width=7cm,height=6.5cm]{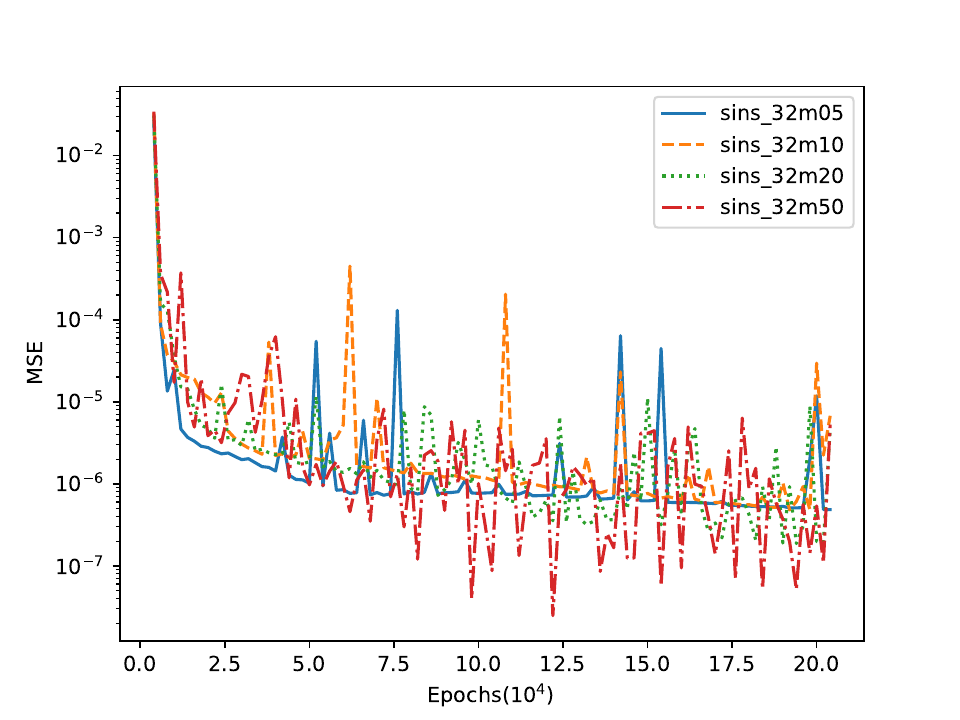}\\
\caption{Training set MSE for different width settings of subnetworks in the approximation of the continuous function. The depth can also influence the approximation rate of TNN networks, especially at a depth of 4.}\label{fig_sins_width}
\end{figure}

In this study, we explored the impact of varying depth and width in neural network architectures on regression tasks. Our approach involved the development of a series of neural networks, each uniquely characterized by their number of layers (depth) and nodes per layer (width). These networks were uniformly denoted as $sins\_depth\_m\_width$, with the structure $sins\_04m05$ exemplifying a TNN with subnetworks composed of 4 layers, each containing 5 nodes. We have different configurations of network width, varying from 5 to 50, and  different network depths, ranging from 4 to 32 layers, as shown in Fig.  \ref{fig_sins_depth} and Fig. \ref{fig_sins_width}. The primary goal was to assess their performance in terms of mean squared error (MSE) on the regression task.

In all cases, there is a rapid decrease in MSE in the early stages of training, indicating that neural networks can quickly capture the main features of the function. As training progresses, the rate of decrease in MSE slows, possibly because the network begins to learn more subtle features of the data. These test results suggest that for the given regression task, the performance of neural networks is influenced by both depth and width, with particular configurations showing varying degrees of effectiveness and stability.

With constant width, an increase in network depth significantly improved early training convergence and, to some extent, enhanced the final MSE. At widths of 20 and 50, there is less variation in convergence speed and final performance among networks of different depths, suggesting that once network width is sufficiently large, increasing depth provides limited performance enhancement. Additionally, this increase in depth moderately improved the final MSE values, suggesting that deeper networks are better at capturing complex patterns in the data, at least up to a certain depth.

With constant depth, an increase in network width led to faster convergence in the initial stages, but did not significantly enhance long-term performance. Across all configurations, deeper networks (such as those with 32 layers) exhibited greater fluctuations during training, likely due to the increased complexity of the network architecture leading to optimization challenges.

To optimize neural networks for approximating unknown functions, fine-tuning of hyperparameters and extensive optimization are typically required. Each network type demands specific hyperparameter settings for optimal performance on the same function.

In our performance comparison, the results are summarized in Table \ref{table:com-sins} for reference.

\begin{table}[H]
  \centering
  \caption{Comparison of Model Performance}
  \label{table:com-sins}
 \begin{tabular}{lccc}
 \toprule  
 Model	& Training MSE & Validation MSE & Testing MSE \\
 \midrule  
 FFN		&	1.7490E-03	& 1.7235E-03	&  2.1336E-03 	\\
 RBN	&	2.1390E-04	& 1.9976E-03	&  1.8986E-03 \\
 TNN 	&	2.9916E-05	& 3.8506E-05	&  4.3672E-05 \\
 \bottomrule  
 \end{tabular}
\end{table}

The TNN demonstrated outstanding performance, consistently yielding the lowest MSE across all measurements. With values ranging from 2.9916E-05 to 4.3672E-05, the TNN not only exhibited high accuracy but also remarkable stability in its predictive capability. This underscores the TNN's efficacy in handling regression tasks, likely attributable to its unique architectural advantages.
In contrast, the FFN, although capable of reaching convergence accuracy comparable to the TNN, displayed notable instability. This variability in performance, as evidenced by MSE values ranging from 1.7490E-03 to 2.1336E-03, suggests that while the FFN has the potential for high precision, achieving consistent outcomes may be challenging. This inconsistency could stem from factors like sensitivity to initial parameter settings or training dynamics, necessitating careful tuning and potentially more sophisticated training methodologies.
The RBN showed mixed results. In one instance, it outperformed the FFN, yet generally, its performance was more aligned with that of the FFN, suggesting a moderate level of effectiveness. With MSE values such as 2.1390E-04 and 1.8986E-03, the RBN appears to be a viable model for certain applications but may not be the most reliable for tasks demanding high precision and stability.

In demonstrating the TNN's proficiency for integration tasks, with the sins function, we employed a TNN featuring a specific subnetwork structure characterized by $L = 3$ layers and $\boldsymbol{p} = (1, 20, 20, 20, 20)$. To further align our training methodology with prior testing environments, aside from the selection of data points, all other training settings for the TNN remained consistent with those used in previous experiments. Specifically, we implemented random sampling to generate 1000 independent and identically distributed (i.i.d.) vectors $\boldsymbol{x}^k, k=1,\ldots, 1000,$ drawn exclusively from the $[0,1]^8$ domain for the sole purpose of TNN training. The results of this comprehensive approach are depicted in Table \ref{table:II-sins} below:

\begin{table}[H]
  \centering
  \caption{Results of Integration Test}
  \label{table:II-sins}
 \begin{tabular}{lccccccc}
 \toprule  
 Model	& Training MSE & $\int_{\Omega}d\boldsymbol{x}(f(\boldsymbol{x})-\Psi(\boldsymbol{x}))$  & $\int_{\Omega}d\boldsymbol{x}(f^2(\boldsymbol{x})-\Psi^2(\boldsymbol{x}))$ \\
 \midrule  
 sins\_03m20		&	1.6173E-04	&	7.2872E-03	&  1.2848	\\
 \bottomrule  
 \end{tabular}
\end{table}

\subsection{Example of 8-Dimensional Product Function}
In our second example, we delve into a continuous function $f(\boldsymbol{x})$, defined over an 8-dimensional input vector $\boldsymbol{x}$. The function is mathematically expressed as:

\begin{equation*}
f(\boldsymbol{x}) =\prod_{i=1}^8 e^{-x_i^2}, \quad \boldsymbol{x} \in [0,1]^8.
\end{equation*}

To ensure methodological consistency in our study, we adopted the same data sampling strategy and neural network framework as utilized in the first example. This entailed generating 1000 independent and identically distributed (i.i.d.) vectors $\boldsymbol{x}^k, k=1,\ldots, 1000,$ each from the domain $[0,1]^8$. We then processed these vectors through the specified function, forming a new dataset comprising input-output pairs $(\boldsymbol{x}^k, f(\boldsymbol{x}^k)),$ for $k=1,\ldots, 1000$. Adhering to the approach of the initial case study, 800 pairs were allocated to the training set, while the remaining 200 were reserved for testing purposes.

\begin{figure}[H]
\centering
\includegraphics[width=7cm,height=6.5cm]{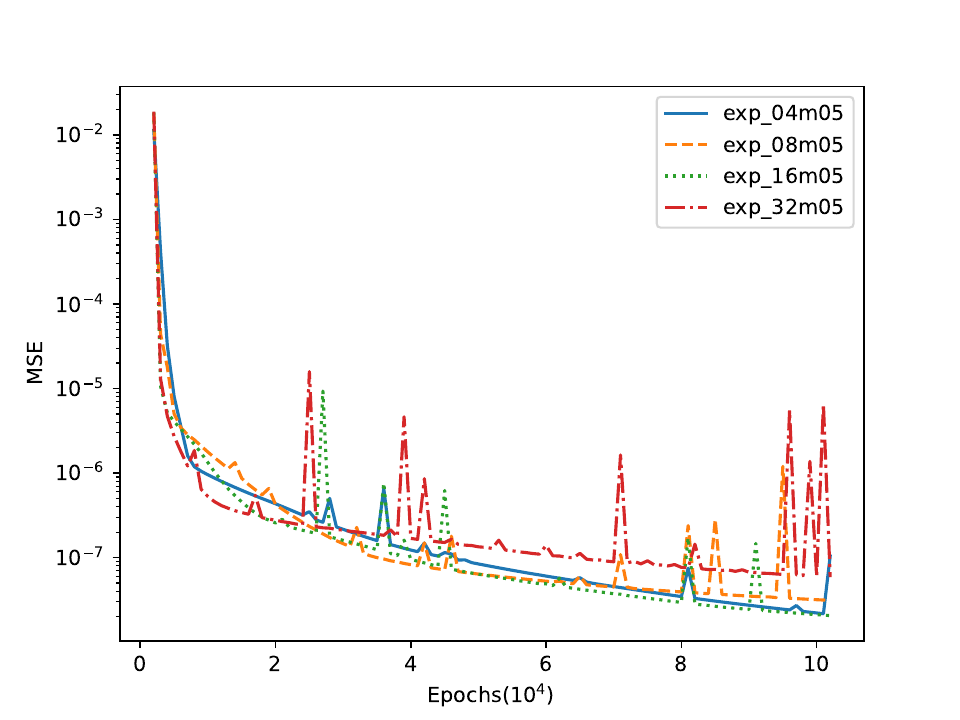}
\includegraphics[width=7cm,height=6.5cm]{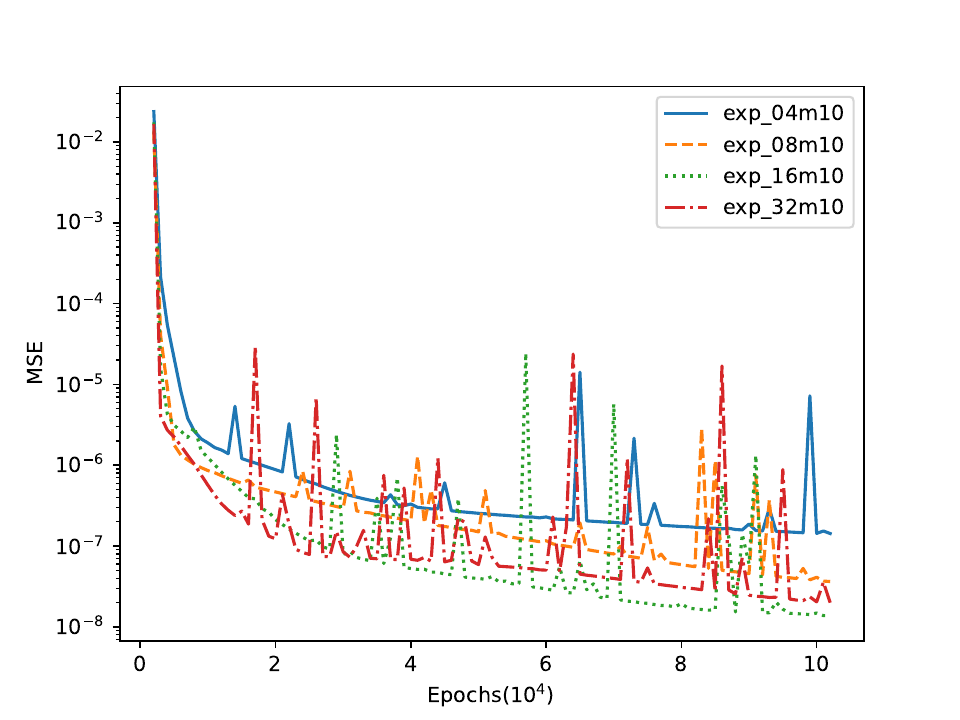}\\
\includegraphics[width=7cm,height=6.5cm]{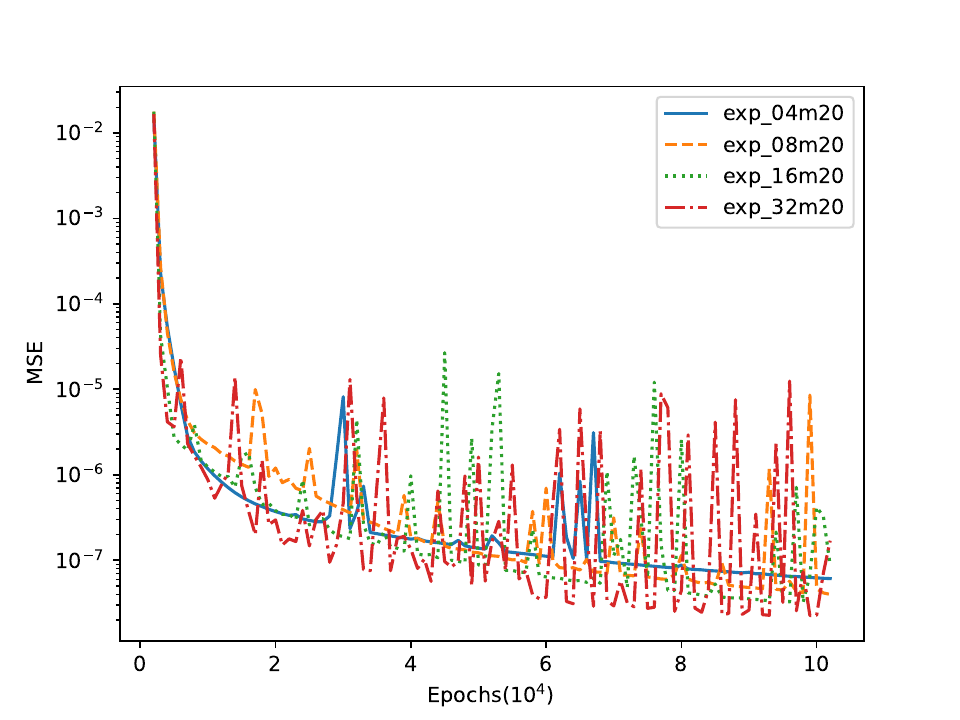}
\includegraphics[width=7cm,height=6.5cm]{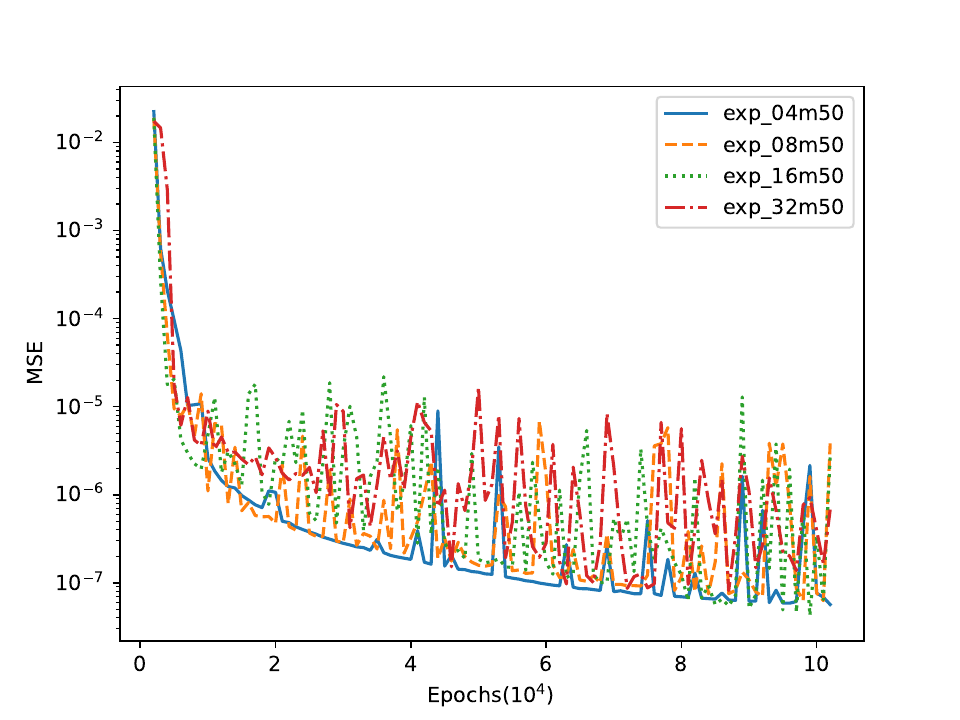}\\
\caption{Training set MSE for different depth setting of subnetworks in the approximation of the continuous function.}\label{fig_exp_depth}
\end{figure}

\begin{figure}[H]
\centering
\includegraphics[width=7cm,height=6.5cm]{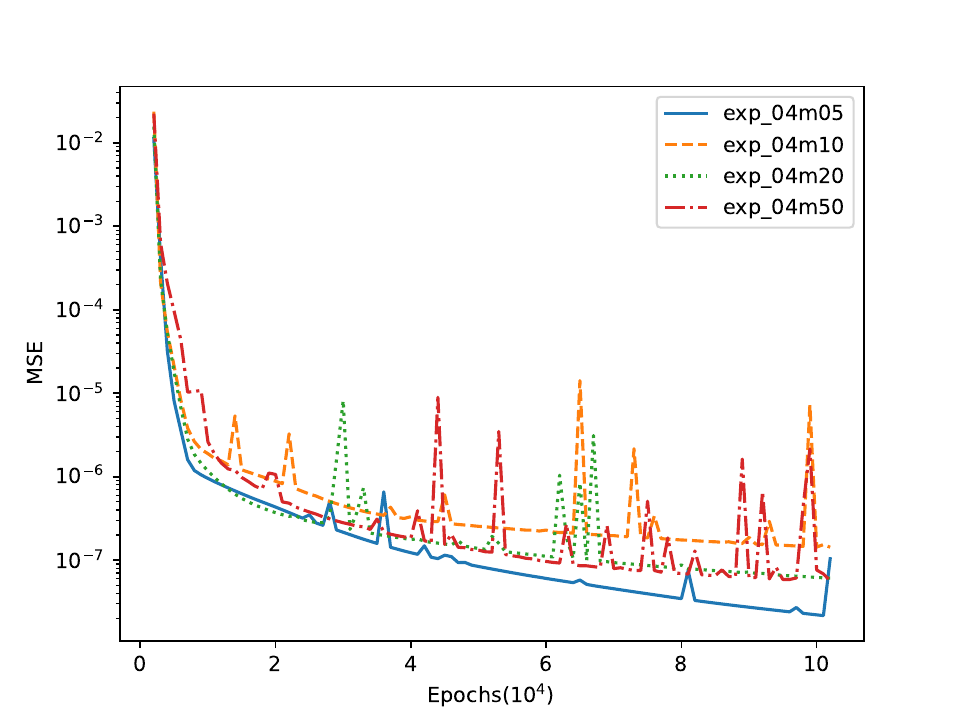}
\includegraphics[width=7cm,height=6.5cm]{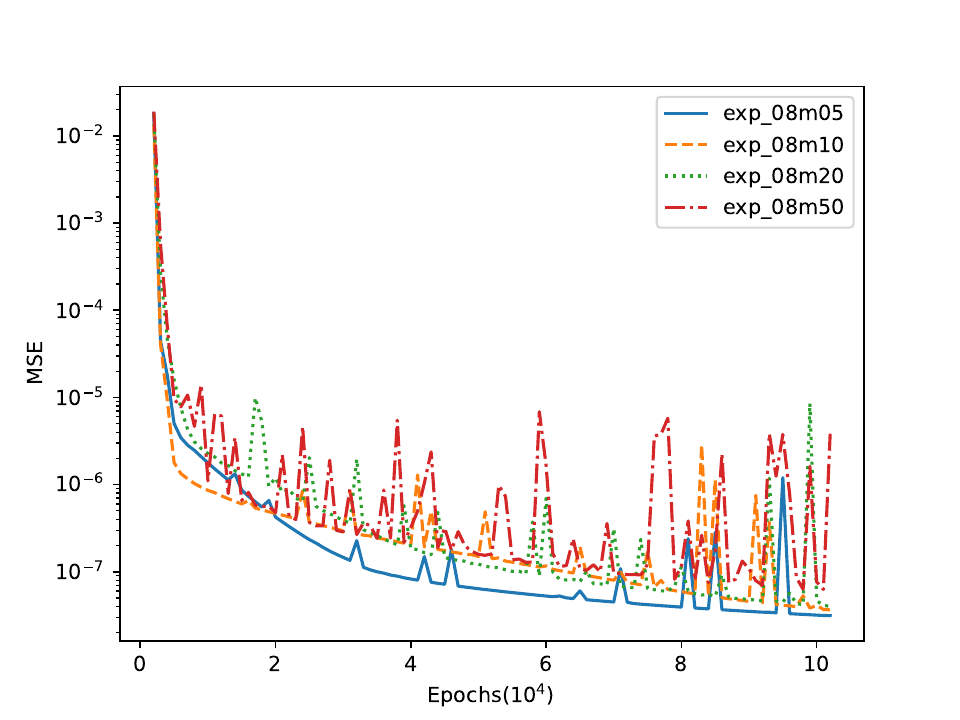}\\
\includegraphics[width=7cm,height=6.5cm]{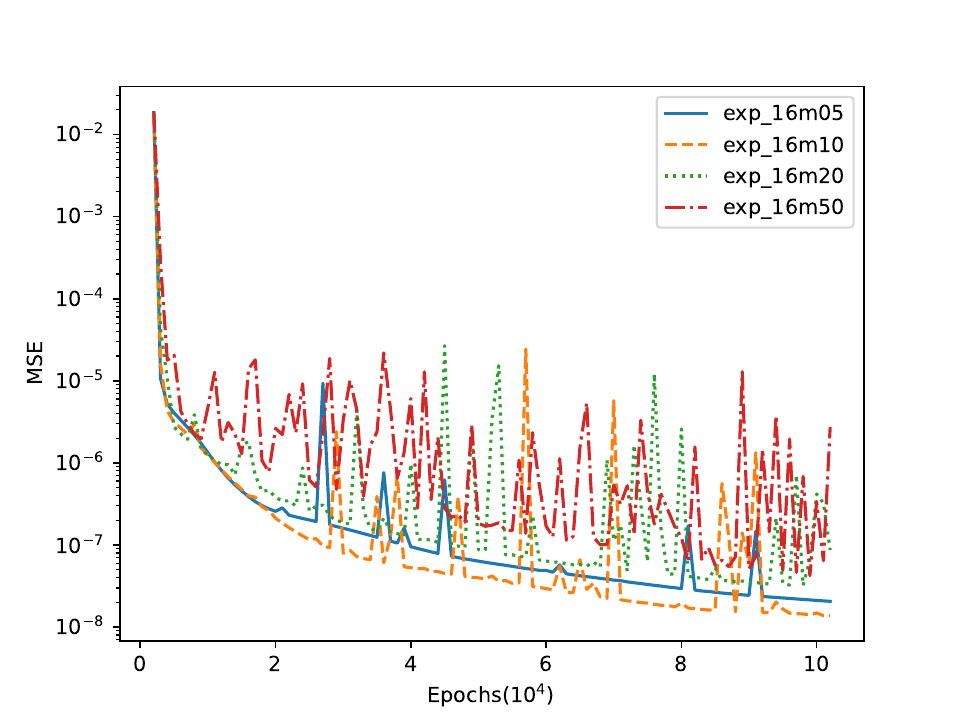}
\includegraphics[width=7cm,height=6.5cm]{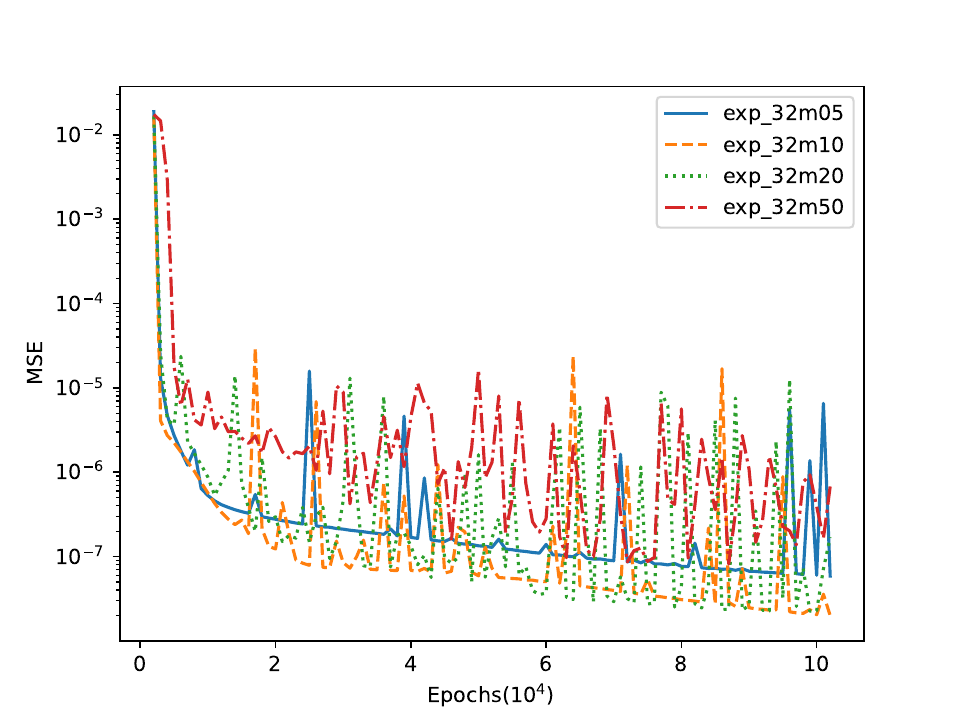}\\
\caption{Training set MSE for different depth setting of subnetworks in the approximation of the continuous function. }\label{fig_exp_width}
\end{figure}

The architecture of the neural networks for this regression task mirrored those implemented in our first example. Denoted as $exp\_depth\_m\_width$, these networks represented a series of TNNs, each with its own distinctive layer (depth) and node (width) configuration. The network structures ranged from $exp\_04m05$, signifying a TNN with 4 layers and 5 nodes per layer, to a variety of other layouts with node widths from 5 to 50 and layer depths from 4 to 32. The Tanh activation function was consistently applied across all network models. Our objective centered on evaluating and comparing the mean squared error (MSE) performance of these networks in relation to the regression challenge introduced by this novel function. The effects of varying network depth and width on the regression performance were systematically illustrated through figures, namely Fig. \ref{fig_exp_width} and Fig. \ref{fig_exp_depth}.

Across all configurations, a steep decline in MSE is observed at the onset of training, indicating that all networks are initially learning the regression task effectively. 

By comparing networks with the same width but different depths (e.g., $exp\_16m05$ vs. $exp\_32m05$), or the same depth but varying widths, it becomes evident that both factors contribute to the learning dynamics, but an optimal balance is crucial. Excessive width or depth can lead to more erratic MSE behavior, while too little limits the network's learning capability. Networks with moderate sizes (like $exp\_16m10$ or $exp\_32m10$) seem to balance learning capacity and generalization better, as evidenced by their smoother convergence curves.

 The graphs also hint at the efficiency of training, where smaller networks reach a performance ceiling with fewer resources, while larger networks continue to improve given more extensive training. However, the diminishing returns on MSE reduction suggest there is a threshold beyond which increasing the network's size yields marginal improvements. This is particularly evident when comparing the performance of large networks with those of moderate size over the same number of epochs.

For this example, we also present a comparative analysis of the models based on their averaged performance metrics. The results, derived from 20 individual runs to ensure statistical robustness, are concisely summarized in Table \ref{table:com-exp}. 

\begin{table}[H]
  \centering
  \caption{Comparison of Model Performance}
  \label{table:com-exp}
 \begin{tabular}{lccc}
 \toprule  
 Model	& Training MSE & Validation MSE & Testing MSE \\
 \midrule  
 FFN		&	3.6227E-06	&  1.3982E-05	&  2.0804E-05 \\
 RBN	&	5.2234E-08	&  9.2906E-07	&  1.3424E-06 \\
 TNN 	&	7.5502E-07	&  7.7727E-07	&  1.2823E-06 \\
 \bottomrule  
 \end{tabular}
\end{table}

In training, the FFN showed a higher MSE (3.6227E-06), suggesting less precise data capture compared to the RBN and TNN. The RBN's extremely low MSE (5.2234E-08) highlighted its fitting capabilities, albeit with potential overfitting concerns. The TNN struck a balance, achieving a MSE of 7.5502E-07.
In the validation phase, the FFN's MSE rose to 1.3982E-05, indicating possible generalization issues. In contrast, the RBN and TNN had lower and similar validation MSEs (9.2906E-07 and 7.7727E-07, respectively), with the TNN slightly outperforming the RBN, suggesting better generalization.
During testing, the FFN's MSE (2.0804E-05) lagged, affirming generalization concerns. The RBN and TNN showed comparable testing performance with MSEs of 1.3424E-06 and 1.2823E-06, respectively, demonstrating their generalization strengths, with a marginal edge for the TNN.


In demonstrating the TNN's proficiency for integration tasks, with the example, we employed a TNN featuring a specific subnetwork structure characterized by $L = 3$ layers and $\boldsymbol{p} = (1, 20, 20, 20, 20)$. To further align our training methodology with prior testing environments, aside from the selection of data points, all other training settings for the TNN remained consistent with those used in previous experiments. Specifically, we implemented random sampling to generate 1000 independent and identically distributed (i.i.d.) vectors $\boldsymbol{x}^k, k=1,\ldots, 1000,$ drawn exclusively from the $[0,1]^8$ domain for the sole purpose of TNN training. The results of this comprehensive approach are depicted in Table \ref{table:II-exp} below:

\begin{table}[H]
  \centering
  \caption{Results of Integration Test}
  \label{table:II-exp}
 \begin{tabular}{lccccccc}
 \toprule  
 Model	& Training MSE & $\int_{\Omega}d\boldsymbol{x}(f(\boldsymbol{x})-\Psi(\boldsymbol{x}))$  & $\int_{\Omega}d\boldsymbol{x}(f^2(\boldsymbol{x})-\Psi^2(\boldsymbol{x}))$ \\
 \midrule  
 exp\_03m20		&	6.9472E-09	&	3.8981E-06	&  4.8862E-06	\\
 \bottomrule  
 \end{tabular}
\end{table}

\subsection{Modeling of Strength of High-Performance Concrete}
The third example in our study shifts focus to the modeling of high-performance concrete strength, a material renowned for its complex and variable composition. High-performance concrete, a critical component in modern construction and infrastructure engineering, is characterized by its high strength, durability, and plasticity. Its formulation typically involves a mix of cement, various aggregates, water, and specialized admixtures, each contributing to its distinctive properties. The versatility of high-performance concrete lies in its ability to be tailored for specific engineering requirements through adjustments in mix proportions and the use of diverse admixtures.

Our dataset for this study comprises 1030 samples, each representing a unique mix of high-performance concrete.  The input vector comprises eight variables: Cement ($kg/m^3$), Fly Ash ($kg/m^3$), Blast Furnace Slag ($kg/m^3$), Water ($kg/m^3$), Superplasticizer ($kg/m^3$), Coarse Aggregate ($kg/m^3$), Fine Aggregate ($kg/m^3$), and Age of Testing ($days$). The output metric is the compressive strength of concrete, measured in MPa. For training purposes, 800 records were selected, with the remaining 230 serving as the test set. 

\begin{figure}[H]
\centering
\includegraphics[width=7cm,height=6.5cm]{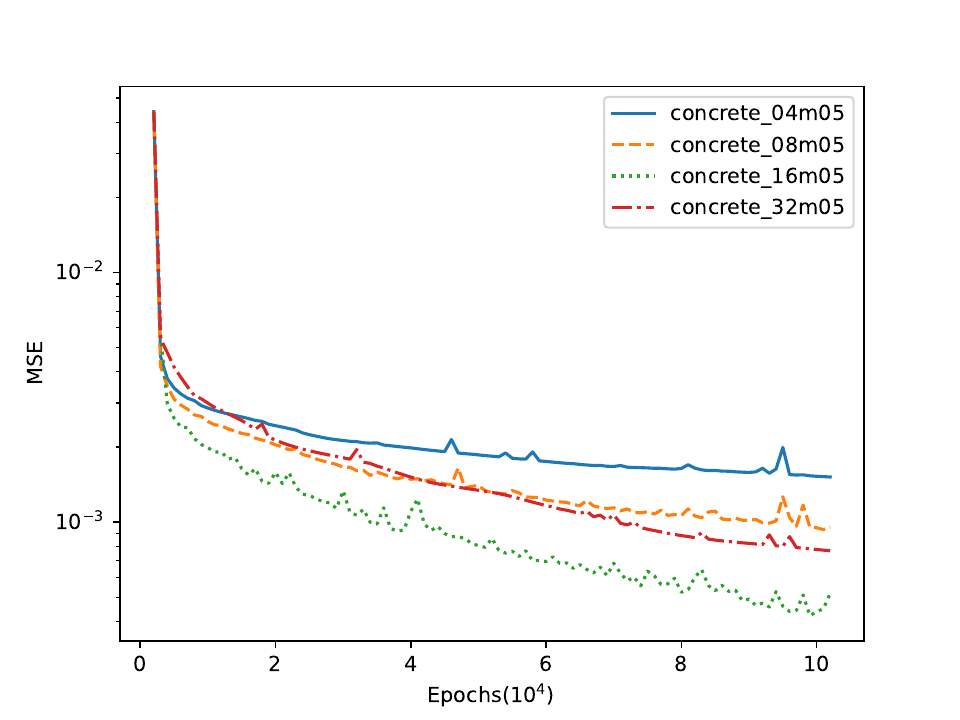}
\includegraphics[width=7cm,height=6.5cm]{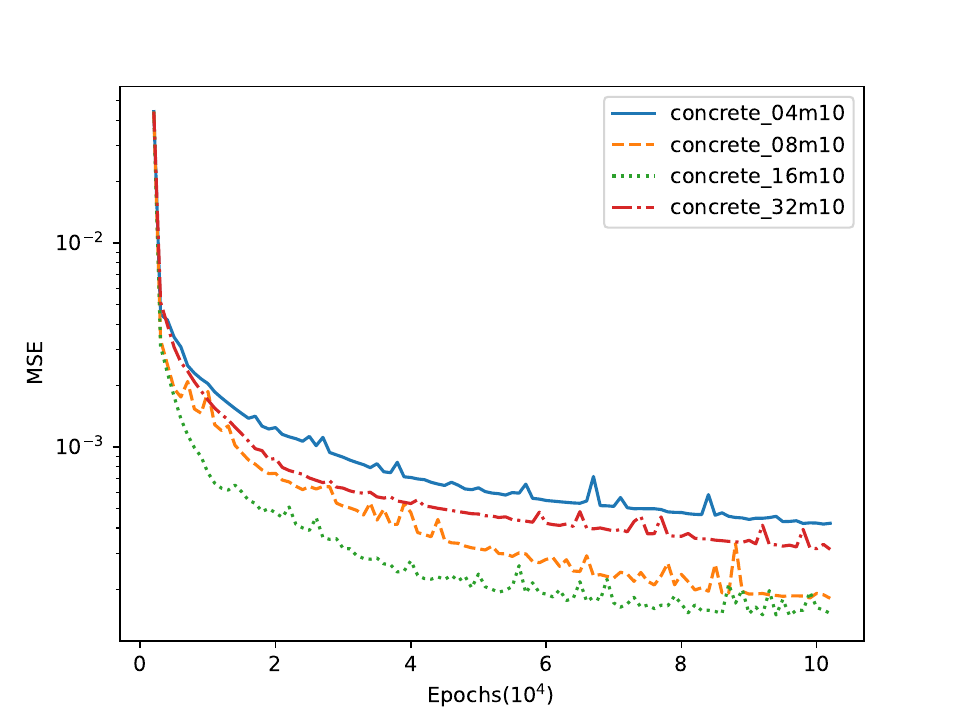}\\
\includegraphics[width=7cm,height=6.5cm]{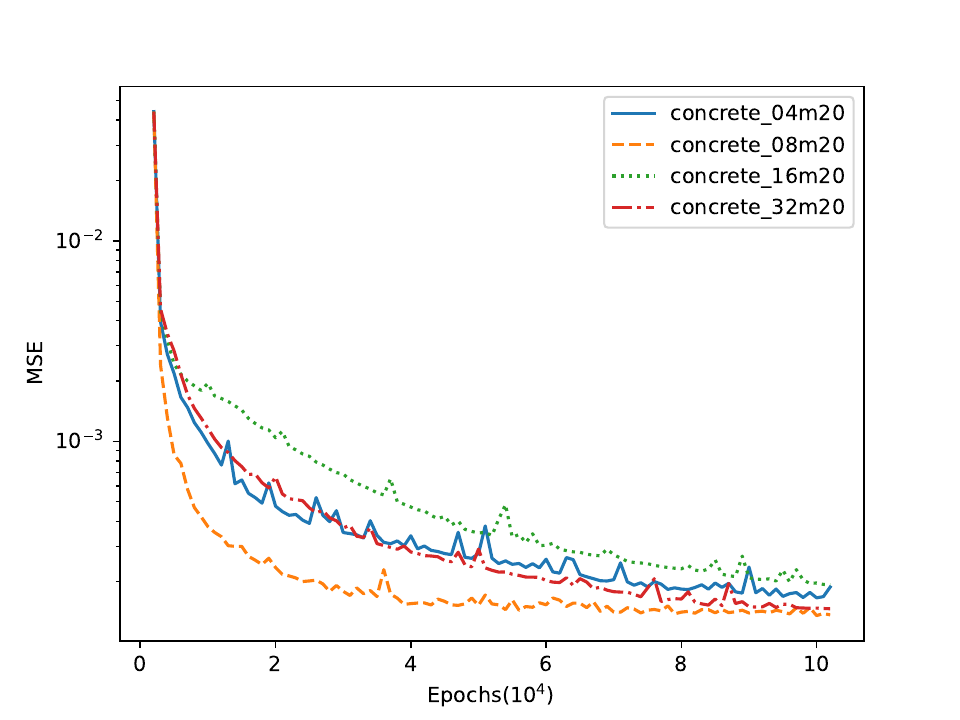}
\includegraphics[width=7cm,height=6.5cm]{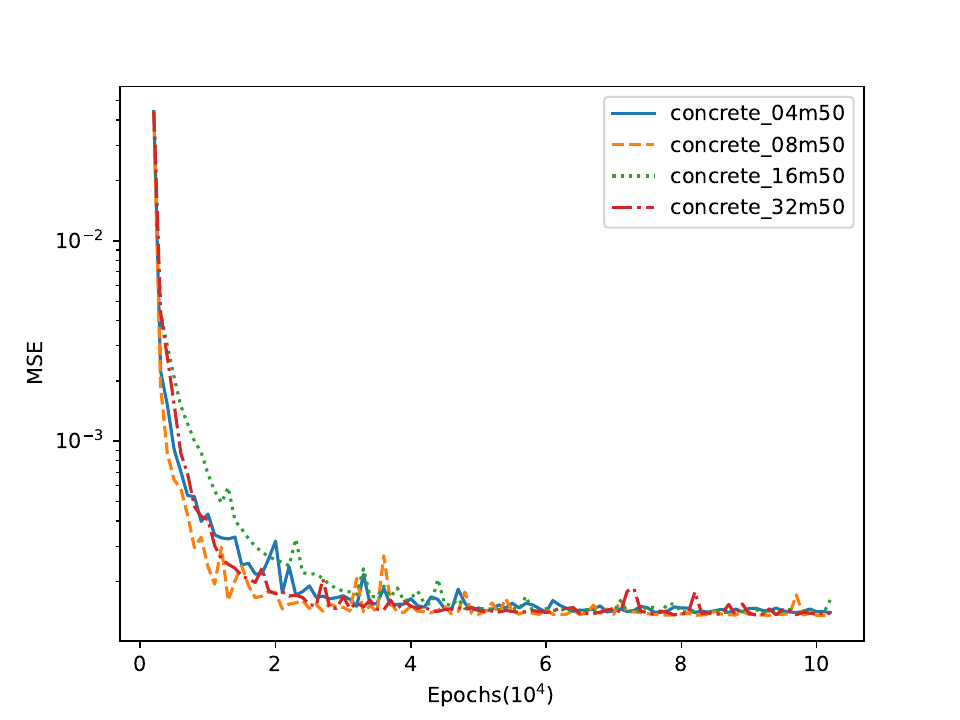}\\
\caption{Training set MSE for different depth setting of subnetworks in the approximation of the continuous function.}\label{fig_concrete_depth}
\end{figure}

\begin{figure}[H]
\centering
\includegraphics[width=7cm,height=6.5cm]{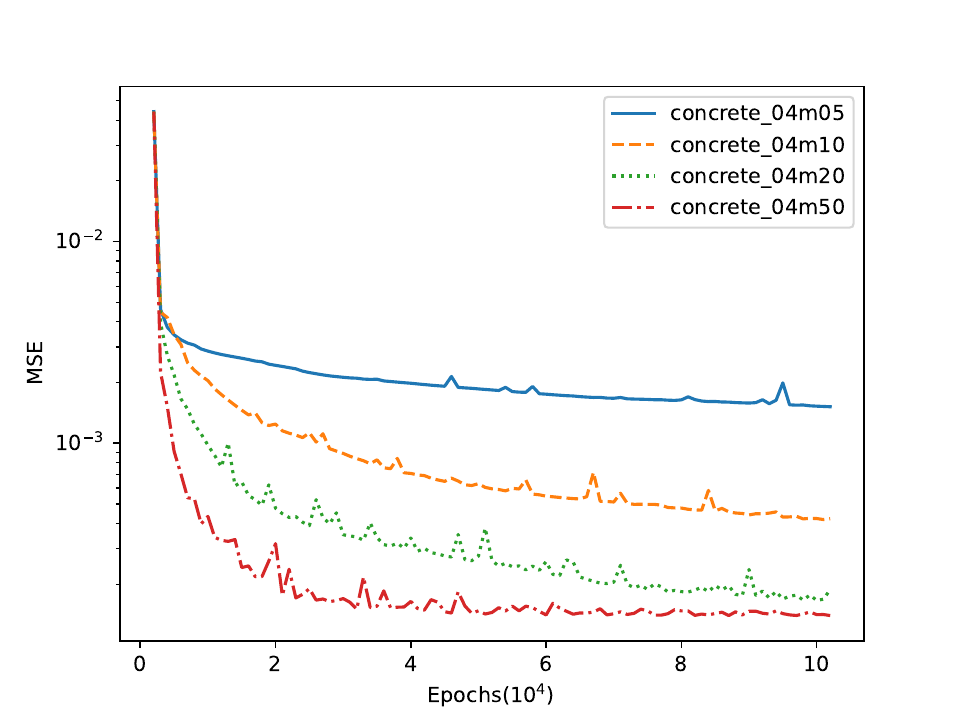}
\includegraphics[width=7cm,height=6.5cm]{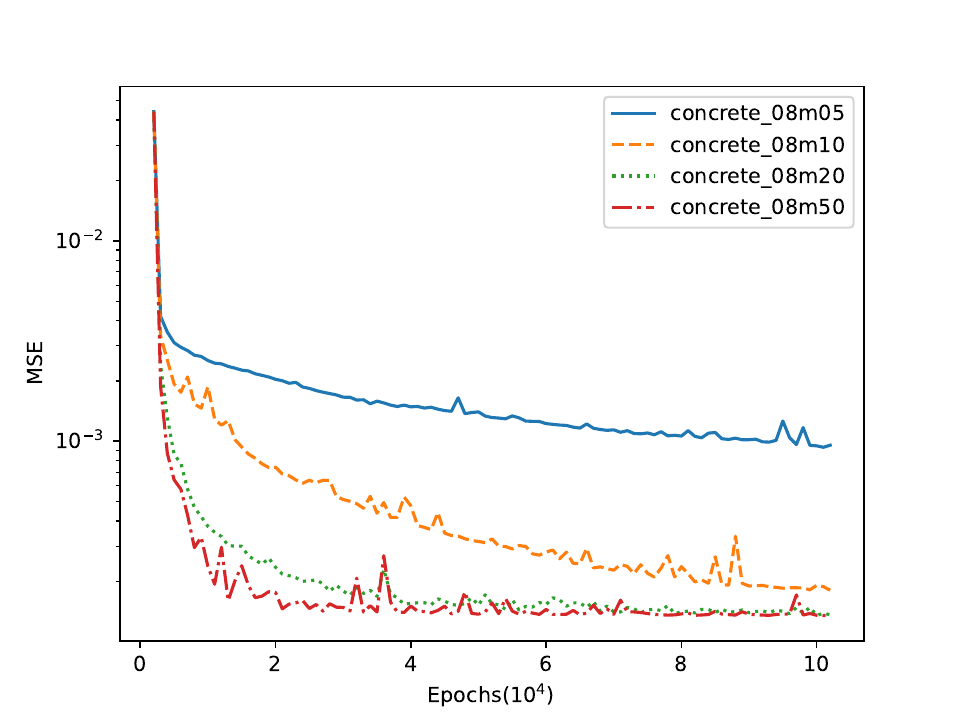}\\
\includegraphics[width=7cm,height=6.5cm]{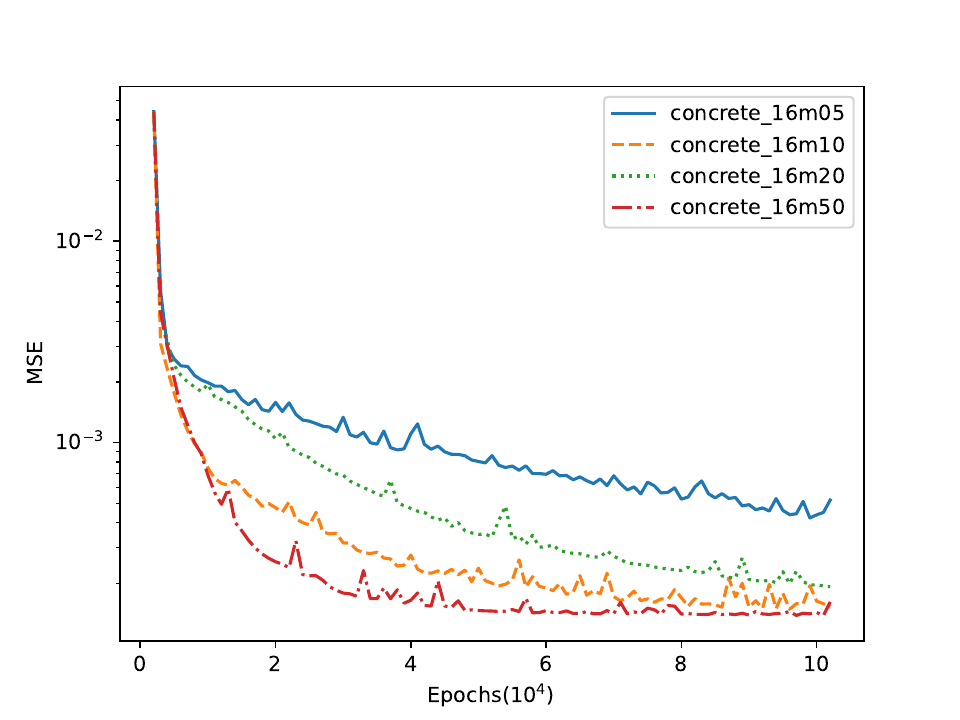}
\includegraphics[width=7cm,height=6.5cm]{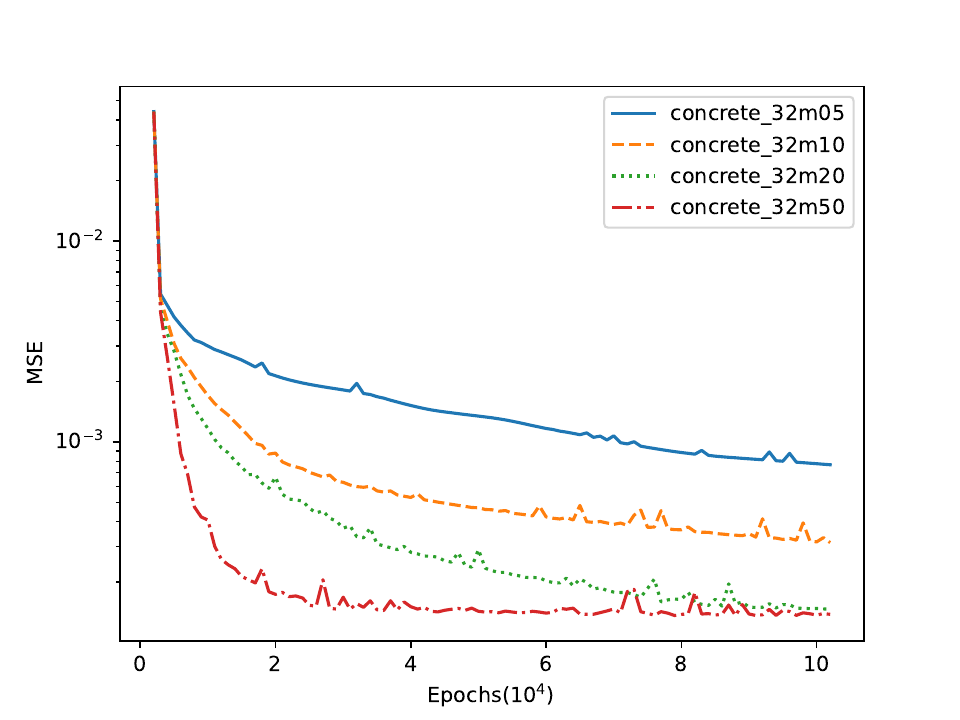}\\
\caption{Training set MSE for different depth setting of subnetworks in the approximation of the continuous function.}\label{fig_concrete_width}
\end{figure}

In this context, our investigation primarily focused on the influence of network architecture, particularly the depth and width of Tensor Neural Networks (TNNs), on their ability to accurately model the concrete's compressive strength. We examined TNNs with varying subnetwork depths (L = 4, 8, 16, 32) and widths (p = 5, 10, 20, 50), assessing their performance in regression tasks. The results of our analysis, particularly focusing on the predictive performance of our model and its ability to accurately estimate the compressive strength, are comprehensively displayed in in Figures \ref{fig_concrete_depth} and \ref{fig_concrete_width}, offering a visual representation of the model's efficacy.

All configurations show a significant decrease in MSE at the beginning of training, with deeper and wider networks (like $concrete\_32m50$) demonstrating a steeper decline. This suggests that networks with higher capacity can quickly capture the underlying relationships in the data.

Comparing networks with similar widths but varying depths, or similar depths with varying widths, suggests that both depth and width contribute to the learning capacity of the network. However, there appears to be a point of diminishing returns, particularly for networks that are significantly wider, such as $concrete\_32m50$, where further increases in size do not equate to proportional decreases in MSE. It is also evident that networks with too much depth relative to their width, such as $concrete\_32m05$, may not perform as well as more balanced configurations, which could be due to difficulties in optimizing deeper networks without sufficient width to support the increased complexity.

In later training epochs, particularly wide networks, exemplified by $concrete\_32m50$, exhibit persistent fluctuations in MSE, hinting at the models' ongoing efforts to minimize training errors, possibly at the expense of overfitting. In contrast, networks of moderate size show a stabilization trend in MSE, indicating a convergence toward solutions with potentially better generalization capabilities.

In summary, the key takeaway from these graphs is the importance of choosing a network configuration that is well-suited to the complexity of the task and the amount of training data available. While larger networks may have the advantage of rapid learning initially, they are also at a higher risk of overfitting. Moderate-sized networks appear to offer a compromise, potentially leading to better generalization without excessive computational cost. 

The comparative analysis of the performances of FFN, RBN and TNN is summarized in Table \ref{table:com-concrete} for reference.

\begin{table}[H]
  \centering
  \caption{Comparison of Model Performance}
  \label{table:com-concrete}
\begin{tabular}{lccc}
\toprule  
 Model	& Training MSE & Validation &	Testing MSE \\
\midrule  
FFN	&	2.7257E-03	& 3.4411E-03	&  3.9812E-03 \\
RBN	&	1.7196E-03	& 4.2468E-03	&  5.1297E-03 	\\
TNN	&	3.1514E-03	& 2.7061E-03	&  3.4946E-03 \\
\bottomrule  
\end{tabular}
\end{table}

The FFN displayed a steady rise in MSE from training (2.7257E-03) to testing (3.9812E-03), reflecting adequate learning during training but reduced effectiveness on new data. This trend suggests a stable yet limited capacity for generalization. Contrastingly, the RBN commenced with the lowest training MSE (1.7196E-03) but experienced a significant performance drop in validation (4.2468E-03) and testing (5.1297E-03), indicating overfitting issues and questionable generalization. Conversely, the TNN presented a reverse trend, beginning with a higher training MSE (3.1514E-03) but outperforming in validation (2.7061E-03) and testing (3.4946E-03). This demonstrates the TNN's ability to generalize more effectively, balancing learning and robustness across datasets.

\begin{figure}[H]
\centering
\includegraphics[width=7cm,height=6.5cm]{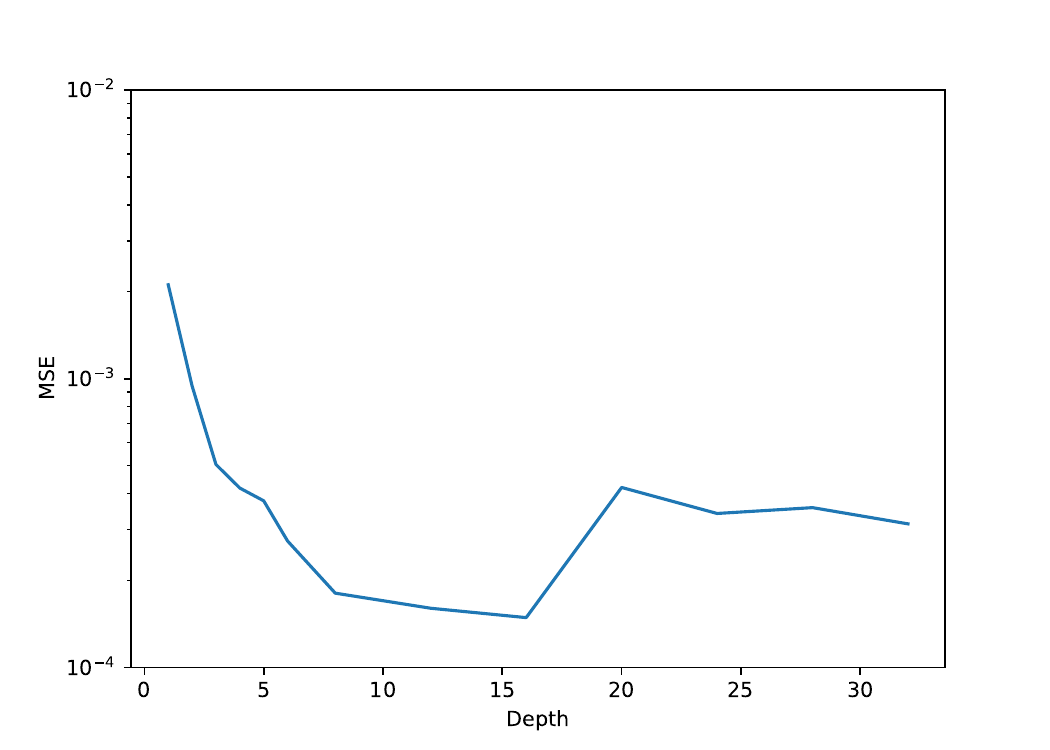}
\includegraphics[width=7cm,height=6.5cm]{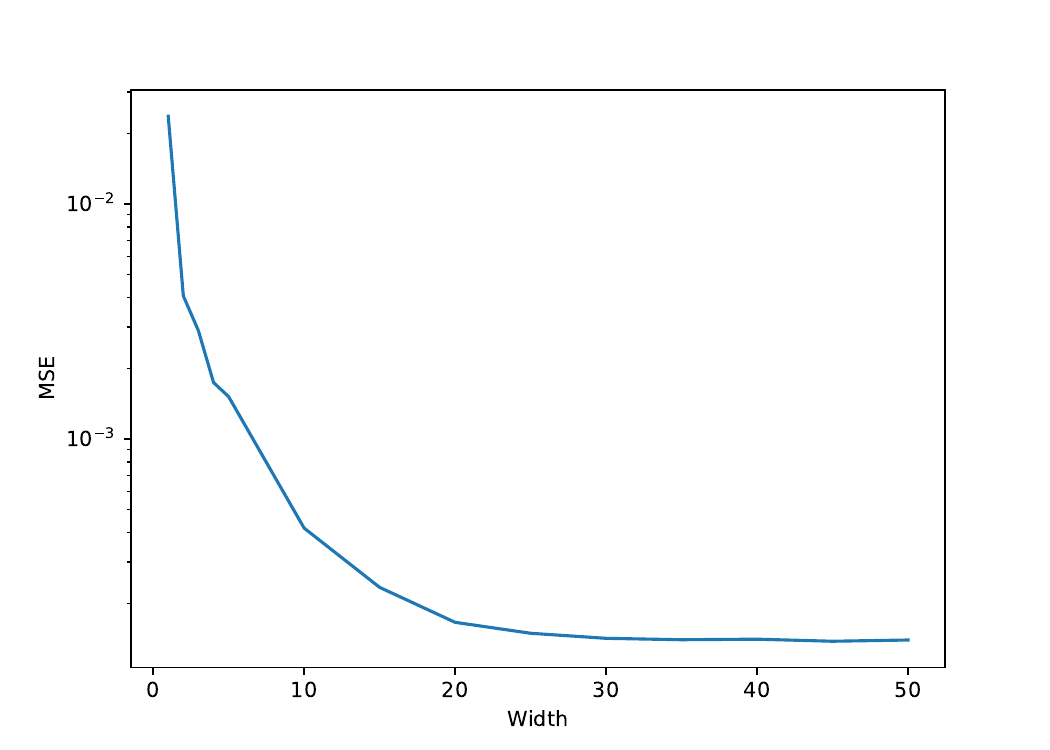}\\
\caption{Minimum Achievable MSE Across Different Network Depths and Widths. As the network depth increases, the MSE demonstrates a trend, which may indicate how deeper networks contribute to the approximation accuracy. This analysis helps in understanding the optimal depth needed to achieve the best performance for the given regression task. The MSE values reflect the network's ability to capture the underlying patterns in the data, with wider networks potentially offering better generalization by accommodating more complex functions.}
\label{fig:depwid}
\end{figure}

Here, we further examine the performance of TNN in regression analysis on real-world data by evaluating the minimum achievable Mean Squared Error (MSE) under varying network widths and depths. Specifically, we compared a series of TNNs with a fixed width of 10 and varying depths, as well as TNNs with a fixed depth of 4 and varying widths. All models were trained for 100,000 iterations using the Adam optimizer. The results, presented in Figure \ref{fig:depwid}, highlight how network depth and width influence the MSE achievable under these conditions.

The analysis of the MSE across different network depths and widths reveals critical insights into the impact of these parameters on the model's performance. As the network depth increases, the MSE shows a trend that suggests deeper networks may enhance approximation accuracy. This trend indicates that increasing depth allows the network to capture more intricate patterns in the data, which could lead to improved performance. However, the results also suggest that there is an optimal depth beyond which the benefits may plateau or even diminish. Similarly, the analysis of varying network widths shows that wider networks generally result in lower MSE values, reflecting the network's enhanced capacity to generalize by fitting more complex functions. This suggests that increasing the width of the network can effectively accommodate more complex patterns in the data, thereby reducing the MSE.

\begin{figure}[H]
\centering
\includegraphics[width=18cm,height=6cm]{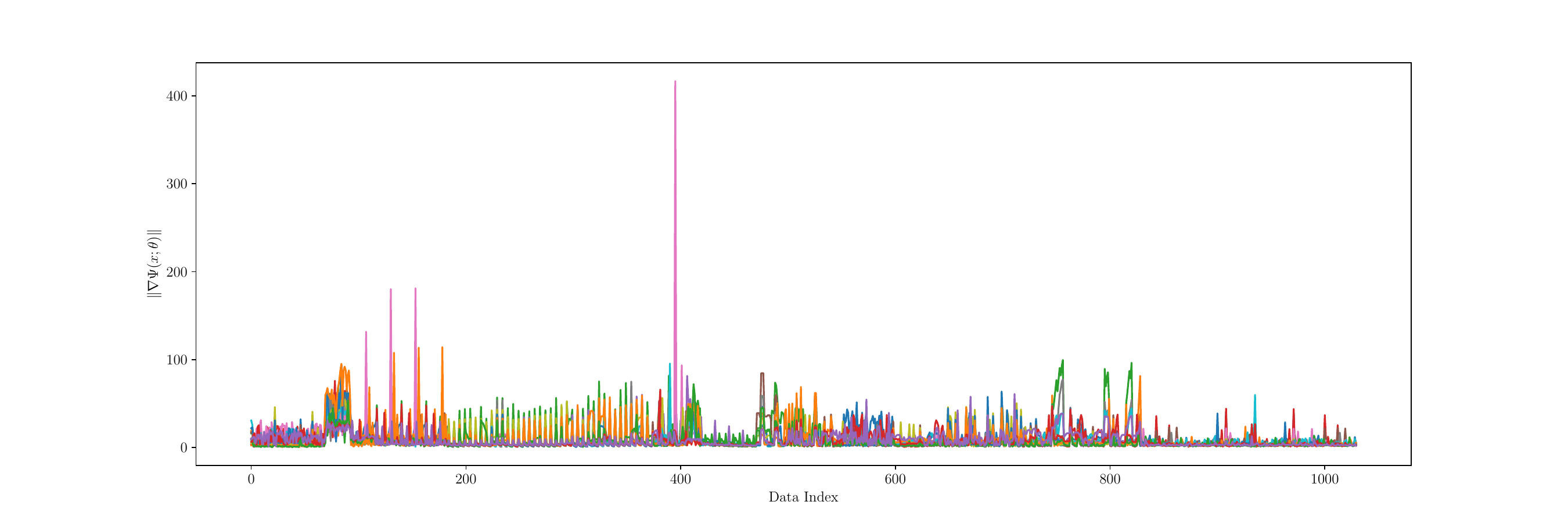}
\caption{Gradient Norm Data Analysis. This figure displays the gradient norms calculated across different data points, showing the overall sensitivity of the model's predictions to changes in the input features. Peaks in the graph represent data points where the model is highly sensitive, indicating regions where the predictions might be more volatile. This information can help in assessing the stability of the model and in identifying areas where the model's predictions may need further refinement.}
\label{fig:gradnorm}
\end{figure}

To investigate the sensitivity of high-strength concrete composition to changes in its constituent components, we utilized the `conctrete\_04m05` model. This model was trained 15 times, each with 200,000 iterations using the Adam optimizer, to investigate the sensitivity of high-strength concrete composition to changes in its constituent components. For each trained model, the Euclidean norm of the gradient \(\|\nabla \Psi(\boldsymbol{x}; \theta)\|\) was calculated at each data point. This norm serves as a metric to quantify the response of the regression function to variations in the input variables.

The resulting data, depicted in Figure \ref{fig:gradnorm}, reveals significant peaks at certain data points, such as around the 100th and 800th indices. These peaks indicate that the regression function is particularly sensitive to changes in specific input variables at these points, which could substantially impact the predicted concrete strength. Meanwhile, the gradient norm is relatively moderate across most data points, suggesting stable sensitivity to input variations in general.

While the Euclidean norm provides a broad overview of the model's sensitivity to input variables, it is insufficient to identify specific dimensions for targeted experimentation. To effectively design future experiments, it is necessary to analyze the individual gradient components. By combining this detailed gradient analysis with Laplacian values \(\Delta \Psi(\boldsymbol{x}; \theta)\), which measure the curvature of the response surface, researchers can pinpoint critical variables. This combination allows for a more refined focus on dimensions where the model is both highly sensitive and potentially unstable.

Focusing on dimensions corresponding to high gradient norms and significant Laplacian values will enable more effective experimental design. By prioritizing these influential variables, subsequent studies can optimize material properties by systematically investigating and adjusting critical components in high-strength concrete composition. This approach ensures that experiments not only explore the most impactful dimensions but also consider the stability and reliability of the model's predictions.

In practical scenarios, when statistical information $\rho_{\boldsymbol{x_{0}}}(\boldsymbol{x})$ regarding specific dimensions of measurement outcomes is available, we can estimate the corresponding value of $f(\boldsymbol{x_{0}})$ through Equations (\ref{Int_TNN_G_f}) and (\ref{Int_TNN2_G_f}). This approach allows us to incorporate known distributions or patterns within the data into our predictions, enhancing the accuracy and relevance of the TNN's output for practical applications.

\section{Conclusion}

This research underscores the unique strengths of Tensor Neural Networks (TNNs) in achieving a balance between precision and generalization, especially notable in regression tasks and predictive modeling. Additionally, TNNs demonstrate a remarkable ability to execute high-dimensional integrations for accurate predictions.

\textbf{Balancing Accuracy and Generalization}: TNNs stand out for their adept balance between achieving high accuracy and excellent generalization in complex regression and predictive tasks. This balance is critical in scenarios where the ability to adapt and perform consistently across diverse data variations is as crucial as reaching high levels of precision. TNNs showcase this through stable performance metrics, such as mean squared error (MSE), across varied training, validation, and testing phases.

\textbf{Predictive Modeling Enhanced by High-Dimensional Integration}: One of the salient advantages of TNNs is their proficiency in efficiently performing high-dimensional integrations. This feature is vital in predictive modeling, particularly when dealing with intricate datasets where understanding the interactions among multiple variables is key. Through high-dimensional integration, TNNs can derive deeper insights and provide more accurate predictions, making them highly suitable for tasks requiring extensive analysis of high-dimensional data.

\textbf{Practical Applications and Future Directions}: The combination of precision and generalization in TNNs, along with their capability for high-dimensional integration, paves the way for their use in fields that demand both accuracy and adaptability. This includes financial analysis, risk modeling, environmental forecasting, and intricate scientific data interpretation. Moreover, the ability to extract detailed information from TNNs through gradient and Laplacian analysis enables the design of more targeted and effective experiments. By focusing on regions with high gradient norms and significant Laplacian values, researchers can prioritize influential variables and explore critical areas more thoroughly. Future research could concentrate on refining these aspects of TNNs, expanding their application to a broader range of domains, and further amplifying their predictive power to address more complex and high-dimensional challenges. 

In conclusion, TNNs' distinctive ability to balance accuracy with generalization, complemented by their skill in high-dimensional integration, solidifies their role as a vital instrument in regression and predictive analytics. This study not only validates the efficacy of TNNs in these areas but also highlights their potential as an indispensable asset in data-driven decision-making across various industries.

\bibliographystyle{unsrt}
\bibliography{tnn4reg}

\begin{thebibliography}{10}

\bibitem{Hastie2009}
Trevor Hastie, Tibshirani Robert, and J.~H. Friedman.
\newblock {\em The Elements of Statistical Learning: Data Mining, Inference,
  and Prediction}.
\newblock Springer, New York, 2009.

\bibitem{Draper1998}
Harry~Smith Norman R.~Draper.
\newblock {\em Applied Regression Analysis}.
\newblock Wiley Series in Probability and Statistics. John Wiley \& Sons, Inc.,
  1998.

\bibitem{montgomery2021}
Douglas~C Montgomery, Elizabeth~A Peck, and G~Geoffrey Vining.
\newblock {\em Introduction to linear regression analysis}.
\newblock John Wiley \& Sons, 2021.

\bibitem{Rosenblatt1958}
F.~Rosenblatt.
\newblock The perceptron: A probabilistic model for information storage and
  organization in the brain.
\newblock {\em Psychological Review}, 65(6):386–408, 1958.

\bibitem{schmidhuber2022}
Juergen Schmidhuber.
\newblock Annotated history of modern ai and deep learning, 2022.
\newblock arXiv:2212.11279.

\bibitem{Nakkiran2020}
Preetum Nakkiran, Gal Kaplun, Yamini Bansal, Tristan Yang, Boaz Barak, and Ilya
  Sutskever.
\newblock Deep double descent: Where bigger models and more data hurt.
\newblock In {\em International Conference on Learning Representations}, 2020.

\bibitem{kaplan2020}
Jared Kaplan, Sam McCandlish, Tom Henighan, Tom~B. Brown, Benjamin Chess, Rewon
  Child, Scott Gray, Alec Radford, Jeffrey Wu, and Dario Amodei.
\newblock Scaling laws for neural language models, 2020.
\newblock arXiv:2001.08361.

\bibitem{lepikhin2020}
Dmitry Lepikhin, HyoukJoong Lee, Yuanzhong Xu, Dehao Chen, Orhan Firat, Yanping
  Huang, Maxim Krikun, Noam Shazeer, and Zhifeng Chen.
\newblock Gshard: Scaling giant models with conditional computation and
  automatic sharding, 2020.
\newblock arXiv:2006.16668.

\bibitem{zhang2021}
Chiyuan Zhang, Samy Bengio, Moritz Hardt, Benjamin Recht, and Oriol Vinyals.
\newblock Understanding deep learning (still) requires rethinking
  generalization.
\newblock {\em Communications of the ACM}, 64(3):107--115, 2021.

\bibitem{Belkin2019}
Mikhail Belkin, Daniel Hsu, Siyuan Ma, and Soumik Mandal.
\newblock Reconciling modern machine-learning practice and the classical
  bias–variance trade-off.
\newblock {\em Proceedings of the National Academy of Sciences},
  116(32):15849--15854, 2019.

\bibitem{Vapnik1998}
Vladimir~N. Vapnik.
\newblock {\em Statistical Learning Theory}.
\newblock Wiley-Interscience, New Jersey, 1998.

\bibitem{Hitchcock1927}
Frank~L. Hitchcock.
\newblock The expression of a tensor or a polyadic as a sum of products.
\newblock {\em Journal of Mathematics and Physics}, 6(1-4):164--189, 1927.

\bibitem{Carroll1970}
J.~Douglas Carroll and Jih-Jie Chang.
\newblock Analysis of individual differences in multidimensional scaling via an
  n-way generalization of ``eckart-young''decomposition.
\newblock {\em Psychometrika}, 35(3):283--319, 1970.

\bibitem{wang2023tensor}
Yifan Wang, Pengzhan Jin, and Hehu Xie.
\newblock Tensor neural network and its numerical integration, 2023.
\newblock arXiv:2207.02754.

\bibitem{li2024tensor}
Yongxin Li, Zhongshuo Lin, Yifan Wang, and Hehu Xie.
\newblock Tensor neural network interpolation and its applications, 2024.
\newblock arXiv:2207.02754.

\bibitem{Györfi2002}
László Györfi, Michael Kohler, Adam Krzyżak, and Harro Walk.
\newblock {\em A Distribution-Free Theory of Nonparametric Regression}.
\newblock Springer Series in Statistics. Springer New York, NY, 2002.

\bibitem{Alexandre2008}
Alexandre~B. Tsybakov.
\newblock {\em Introduction to Nonparametric Estimation}.
\newblock Springer Series in Statistics. Springer New York, NY, 2008.

\bibitem{KURKOVA1992501}
Věra Kůrková.
\newblock Kolmogorov's theorem and multilayer neural networks.
\newblock {\em Neural Networks}, 5(3):501--506, 1992.

\bibitem{Chen1995}
Tianping Chen and Hong Chen.
\newblock Universal approximation to nonlinear operators by neural networks
  with arbitrary activation functions and its application to dynamical systems.
\newblock {\em IEEE Transactions on Neural Networks}, 6(4):911--917, 1995.

\bibitem{Hieber2020}
Johannes Schmidt-Hieber.
\newblock {Nonparametric regression using deep neural networks with ReLU
  activation function}.
\newblock {\em The Annals of Statistics}, 48(4):1875 -- 1897, 2020.

\bibitem{Rumelhart1988}
David~E. Rumelhart, Geoffrey~E. Hinton, and Ronald~J. Williams.
\newblock {\em Learning Representations by Back-Propagating Errors}, page
  696–699.
\newblock MIT Press, Cambridge, MA, USA, 1988.

\bibitem{broomhead1988}
D.S. Broomhead, David Lowe, ROYAL SIGNALS, and RADAR ESTABLISHMENT
  MALVERN~(UNITED KINGDOM).
\newblock Radial basis functions, multi-variable functional interpolation and
  adaptive networks.
\newblock Technical report, 1988.

\bibitem{Huang2003}
Guang-Bin Huang.
\newblock Learning capability and storage capacity of two-hidden-layer
  feedforward networks.
\newblock {\em IEEE Transactions on Neural Networks}, 14(2):274--281, 2003.

\bibitem{reed1999}
Russell Reed and Robert~J MarksII.
\newblock {\em Neural smithing: supervised learning in feedforward artificial
  neural networks}.
\newblock Mit Press, 1999.

\bibitem{looney1997}
Carl~Grant Looney.
\newblock {\em Pattern recognition using neural networks: theory and algorithms
  for engineers and scientists}.
\newblock Oxford University Press, Inc., 1997.

\bibitem{Bianchini1995}
M.~Bianchini, P.~Frasconi, and M.~Gori.
\newblock Learning without local minima in radial basis function networks.
\newblock {\em IEEE Transactions on Neural Networks}, 6(3):749--756, 1995.

\bibitem{Poggio1990}
T.~Poggio and F.~Girosi.
\newblock Networks for approximation and learning.
\newblock {\em Proceedings of the IEEE}, 78(9):1481--1497, 1990.

\bibitem{Haykin2007}
Simon Haykin.
\newblock {\em Neural Networks: A Comprehensive Foundation (3rd Edition)}.
\newblock Prentice-Hall, Inc., USA, 2007.

\bibitem{maren2014handbook}
Alianna~J Maren, Craig~T Harston, and Robert~M Pap.
\newblock {\em Handbook of neural computing applications}.
\newblock Academic Press, 2014.

\bibitem{haykin2009neural}
Simon~S. Haykin.
\newblock {\em Neural networks and learning machines}.
\newblock Pearson Education, Upper Saddle River, NJ, third edition, 2009.

\bibitem{chen1991}
S.~Chen, C.F.N. Cowan, and P.M. Grant.
\newblock Orthogonal least squares learning algorithm for radial basis function
  networks.
\newblock {\em IEEE Transactions on Neural Networks}, 2(2):302--309, 1991.

\bibitem{Jiang2022}
Qinghua Jiang, Lailai Zhu, Chang Shu, and Vinothkumar Sekar.
\newblock An efficient multilayer rbf neural network and its application to
  regression problems.
\newblock {\em Neural Computing and Applications}, 34(6):4133--4150, 2022.

\bibitem{Chao2001}
J.~Chao, M.~Hoshino, T.~Kitamura, and T.~Masuda.
\newblock A multilayer rbf network and its supervised learning.
\newblock In {\em IJCNN'01. International Joint Conference on Neural Networks.
  Proceedings (Cat. No.01CH37222)}, volume~3, pages 1995--2000 vol.3, 2001.

\bibitem{YEH1998}
I.-C. Yeh.
\newblock Modeling of strength of high-performance concrete using artificial
  neural networks.
\newblock {\em Cement and Concrete Research}, 28(12):1797--1808, 1998.

\bibitem{Eckart1936}
Carl Eckart and Gale Young.
\newblock The approximation of one matrix by another of lower rank.
\newblock {\em Psychometrika}, 1(3):211--218, 1936.

\bibitem{paszke2017automatic}
Adam Paszke, Sam Gross, Soumith Chintala, Gregory Chanan, Edward Yang, Zachary
  DeVito, Zeming Lin, Alban Desmaison, Luca Antiga, and Adam Lerer.
\newblock Automatic differentiation in pytorch.
\newblock In {\em NIPS-W}, 2017.

\bibitem{kingma2017adam}
Diederik~P. Kingma and Jimmy Ba.
\newblock Adam: A method for stochastic optimization, 2017.
\newblock arXiv:1412.6980.

\bibitem{Bottou2010}
L{\'e}on Bottou.
\newblock Large-scale machine learning with stochastic gradient descent.
\newblock In Yves Lechevallier and Gilbert Saporta, editors, {\em Proceedings
  of COMPSTAT'2010}, pages 177--186, Heidelberg, 2010. Physica-Verlag HD.

\bibitem{Yeh2007}
I-Cheng Yeh.
\newblock {Concrete Compressive Strength}.
\newblock UCI Machine Learning Repository, 2007.
\newblock {DOI}: https://doi.org/10.24432/C5PK67.

\bibitem{wasserman1993}
Philip~D. Wasserman.
\newblock {\em Advanced Methods in Neural Computing}.
\newblock John Wiley \& Sons, Inc., USA, 1st edition, 1993.

\end{thebibliography}

\end{document}